\documentclass{article}
\usepackage{amssymb}
\usepackage{amsmath}
\usepackage{amsthm}
\usepackage[letterpaper]{geometry}
\usepackage[round]{natbib}
\usepackage[dvipsnames]{xcolor}
\usepackage[colorlinks,linktoc=true,
            citecolor= Cerulean!85!green!90!black,%Cerulean!75!green, %JungleGreen!70!blue, %CornflowerBlue
            linkcolor=YellowOrange,
            urlcolor=RubineRed!85!black,
            anchorcolor=YellowOrange!85!black
            ]{hyperref}

\usepackage[utf8]{inputenc} % allow utf-8 input
\usepackage[T1]{fontenc}    % use 8-bit T1 fonts
\usepackage{hyperref}       % hyperlinks
\usepackage{url}            % simple URL typesetting
\usepackage{amsfonts}       % blackboard math symbols
\usepackage{nicefrac}       % compact symbols for 1/2, etc.
\usepackage{graphicx}
\usepackage{subcaption}
\usepackage{array}
\usepackage{multirow}
\usepackage{setspace}
\usepackage{booktabs}       % professional-quality tables
\usepackage{framed}
\usepackage{mathtools}
\usepackage{bm}
\usepackage{xspace}
\usepackage{todonotes}

\graphicspath{{./plots/}}

\DeclareMathOperator*{\argmin}{arg\,min}

\newcommand{\shorteqref}[1]{(\ref{#1})}

\usepackage{adjustbox}
\usepackage{rotating}
\usepackage{wrapfig}
\usepackage{enumitem,kantlipsum}

\usepackage{pgfplots}
\usepackage{amsfonts}
\usepackage{tikz}
\usepackage{caption}
\usepackage{bm}
\usepackage{booktabs,arydshln}
\usepackage{readarray}
\usepackage{todonotes}
\usepackage{float}
\usepackage{csvsimple}
\usepackage[normalem]{ulem}
\usepackage{multirow}
\usepackage{textcomp}
\usepackage{gensymb}
\usepackage{wasysym}
\usepackage{oplotsymbl}
\usepackage{ifsym}
\usepackage{relsize}
\usepackage{tabto}
\usepackage{pifont}
\usepackage{makecell}

\pgfplotsset{compat=newest}% to avoid the pgfplots warning
\usetikzlibrary{intersections, pgfplots.fillbetween}
\usetikzlibrary{shapes.geometric,snakes,calc}
\usetikzlibrary{pgfplots.groupplots}
\usetikzlibrary{colorbrewer}
\usetikzlibrary{patterns}
\usepgfplotslibrary{colorbrewer}

\colorlet{exogencolor}{Turquoise}
\colorlet{endogencolor}{BurntOrange}
\colorlet{residualcolor}{BrickRed}
\colorlet{residualcolor2}{Salmon}
\colorlet{poscolor}{Green}
\colorlet{negcolor}{Red}
\colorlet{refcolor}{ProcessBlue}

\usepgfplotslibrary{external} 
\usetikzlibrary{external}
\makeatletter
  {\end{list}
}
\makeatother
\makeatletter

\makeatletter
\def\adl@drawiv#1#2#3{%
        \hskip.5\tabcolsep
        \xleaders#3{#2.5\@tempdimb #1{1}#2.5\@tempdimb}%
                #2\z@ plus1fil minus1fil\relax
        \hskip.5\tabcolsep}
\newcommand{\cdashlinelr}[1]{%
  \noalign{\vskip\aboverulesep
           \global\let\@dashdrawstore\adl@draw
           \global\let\adl@draw\adl@drawiv}
  \cdashline{#1}
  \noalign{\global\let\adl@draw\@dashdrawstore
           \vskip\belowrulesep}}
\makeatother

\newcommand{\basicemb}{\texttt{BasicEmb}}
\newcommand{\residualthresh}{\texttt{ResTresh}}
\newcommand{\twostepmodelregr}{\texttt{ResEmbRegr}}

\newcommand{\domainadv}{\texttt{EnvInvEmb}}
\newcommand{\catchtwotwo}{\texttt{Catch22}}
\newcommand{\residualcatchtwotwo}{\texttt{ResCatch22}}

\newcommand{\localoutlierfactor}{\texttt{LOF}}
\newcommand{\isolationforest}{\texttt{IF}}
\newcommand{\ocsvm}{\texttt{OCSVM}}
\newcommand{\ominanomaly}{\texttt{OmniAnomaly}}

\newcommand{\turbine}{Turbine}
\newcommand{\devops}{DevOps}
\newcommand{\synthetic}{Synthetic}
\newcommand{\pendulum}{Pendulum}

\newcommand{\zts}{\mathbf{z}}
\newcommand{\xts}{\mathbf{x}}
\newcommand{\yts}{\mathbf{y}}
\newcommand{\cts}{\mathbf{c}}
\newcommand{\pos}{\text{pos}}
\newcommand{\nega}{\text{neg}}
\newcommand{\ctx}{\text{ctx}}

\title{
Intrinsic Anomaly Detection for Multi-Variate Time Series
}

\author{%
  Stephan Rabanser\thanks{Equal contribution.} \thanks{Work done while at Amazon Research.} \\
  University of Toronto \& Vector Institute \\
  \texttt{stephan@cs.toronto.edu}\\
  \and
  Tim Januschowski\footnotemark[1] \footnotemark[2]\\
  Zalando Research\\
  \texttt{tim.januschowski@zalando.de}\\
  \and
  Kashif Rasul\\
  Morgan Stanley\\
  \texttt{rasul.kashif@morganstanley.com}\\
  \and
  Oliver Borchert\footnotemark[2]\\
  Technical University of Munich \\
  \texttt{borchero@in.tum.de}\\
  \and
  Richard Kurle\\
  Amazon Research\\
  \texttt{kurler@amazon.com}\\
  \and
  Jan Gasthaus\\
  Amazon Research\\
  \texttt{gasthaus@amazon.com}
  \and
  Michael Bohlke-Schneider\\
  Amazon Research\\
  \texttt{bohlkem@amazon.com}
  \and
  Nicolas Papernot\\
  University of Toronto \& Vector Institute\\
  \texttt{nicolas.papernot@utoronto.ca}\\
  \and
  Valentin Flunkert\\
  Amazon Research\\
  \texttt{flunkert@amazon.com}
}

\date{} 

\begin{document}

\maketitle

\begin{abstract}
    \noindent We introduce a novel, practically relevant variation of the anomaly detection problem in multi-variate time series: \emph{intrinsic anomaly detection}. It appears in diverse practical scenarios ranging from DevOps to IoT, where we want to recognize failures of a system that operates under the influence of a surrounding environment. 
Intrinsic anomalies are changes in the functional dependency structure between time series that represent an environment and time series that represent the internal state of a system that is placed in said environment. 
We formalize this problem, provide under-studied public and new purpose-built data sets for it, and present methods that handle intrinsic anomaly detection. These address the short-coming of existing anomaly detection methods that cannot differentiate between expected changes in the system's state and unexpected ones, i.e., changes in the system that deviate from the environment's influence.   
Our most promising approach is fully unsupervised and combines adversarial learning and time series representation learning, thereby addressing problems such as label sparsity and subjectivity, while allowing to navigate and improve notoriously problematic anomaly detection data sets. 
\end{abstract}

\section{Introduction}
\label{sec:intro}

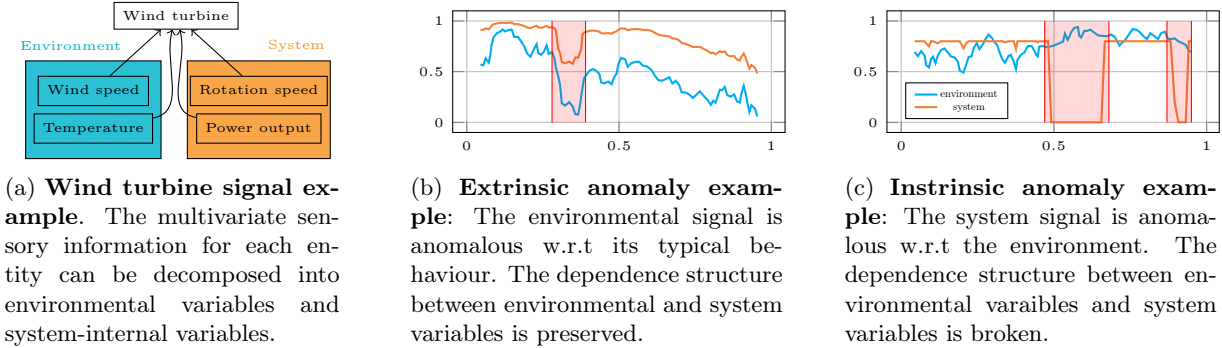
\begin{figure*}
    \centering
    
    \begin{subfigure}[t]{0.27\textwidth}
        \centering
        
        %\begin{adjustbox}{width=\textwidth}
    \begin{tikzpicture}
    \tikzstyle{every node}=[font=\tiny]
    
    \node[draw] (app) at (0,0) {Wind turbine};
    
    \node[exogencolor] (ex) at (-1.4,-0.4) {Environment};
    \node[endogencolor] (end) at (1.6,-0.4) {System};
    
    \draw[draw, fill=exogencolor!80] (-2.0,-1.9) rectangle ++(1.8,1.3);
    \draw[draw, fill=endogencolor!80] (0.15,-1.9) rectangle ++(1.9,1.3);
    %\draw[draw, fill=endogencolor] (0,-2.7) rectangle ++(4.75,1.4);
    
    \node[draw] (rr) at (-1.1,-1) {Wind speed};
    \node[draw] (tr) at (-1.1,-1.5) {Temperature};
    
    \node[draw] (cpu) at (1.1,-1) {Rotation speed};
    \node[draw] (ram) at (1.1,-1.5) {Power output};
    
    % \node[] () at (1.1,-1.95) {$\vdots$};
    % \node[] () at (-1.1,-1.95) {$\vdots$};
    
    % \node[draw] (avg) at (-1.2,-2.35) {avg};
    % \node[draw] (std) at (-0.4,-2.35) {std};
    % \node[draw] (min) at (0.4,-2.35) {min};
    % \node[draw] (max) at (1.2,-2.35) {max};
    
    \draw [->] (rr) to (app);
    \draw [->] (cpu) to (app);
    \draw [->] (tr) to [out=10,in=255] (app);
    \draw [->] (ram) to [out=170,in=285] (app);

    \end{tikzpicture}
    %\end{adjustbox}
    \caption{\textbf{Wind turbine signal example}. The multivariate sensory information for each entity can be decomposed into environmental variables and system-internal variables.}
    \label{fig:sensor_hierarchy}
        
    \end{subfigure}
    \hspace{20pt}
    \begin{subfigure}[t]{0.3\textwidth}
        \centering
        \begin{tikzpicture}
            \path[fill=white] (2.9,-0.35) rectangle ++(1,1);
            \begin{axis}[
              height=3.2cm,
              width=6cm,
              legend style={at={(0.15,0.08)},anchor=south,nodes={scale=0.4, transform shape}},
              grid=major,
              ticklabel style = {font=\tiny},
              label style={font=\tiny},
            ]
            
            \addplot [Cyan, no marks, thick] table [x=x, y=e, col sep=comma] {figs/wt_ext.csv};
            %\addlegendentry{env.}
            \addplot [Orange, no marks, thick] table [x=x, y=i, col sep=comma] {figs/wt_ext.csv};
            %\addlegendentry{sys.}
            \addplot +[mark=none,name path=start,red] coordinates {(0.28, 0) (0.28, 1)};
            \addplot +[mark=none,name path=end,red] coordinates {(0.39, 0) (0.39, 1)};
            \tikzfillbetween[of=start and end]{red, opacity=0.15};
            \end{axis}
        \end{tikzpicture}
        \caption{\textbf{Extrinsic anomaly example}: The environmental signal is anomalous w.r.t its typical behaviour. The dependence structure between environmental and system variables is preserved.}
        \label{fig:extr_anom}
    \end{subfigure}%
    \hspace{20pt}
    \begin{subfigure}[t]{0.3\textwidth}
        \centering
        \begin{tikzpicture}
        
            \pgfplotsset{
            compat=1.11,
            legend image code/.code={
            \draw[mark repeat=2,mark phase=2]
            plot coordinates {
            (0cm,0cm)
            (0.15cm,0cm)        %% default is (0.3cm,0cm)
            (0.3cm,0cm)         %% default is (0.6cm,0cm)
            };%
            }
            }
        
            \path[fill=white] (2.9,-0.35) rectangle ++(1,1);
            \begin{axis}[
              height=3.2cm,
              width=6cm,
              legend style={at={(0.21,0.11)},anchor=south,nodes={scale=0.4, transform shape}},
              grid=major,
              ticklabel style = {font=\tiny},
              label style={font=\tiny},
            ]
            
            \addplot [Cyan, no marks, thick] table [x=x, y=e, col sep=comma] {figs/wt_int.csv};
            \addlegendentry{environment}
            \addplot [Orange, no marks, thick] table [x=x, y=i, col sep=comma] {figs/wt_int.csv};
            \addlegendentry{system}
            \addplot +[mark=none,name path=start,red] coordinates {(0.47, 0) (0.47, 1)};
            \addplot +[mark=none,name path=end,red] coordinates {(0.68, 0) (0.68, 1)};
            \tikzfillbetween[of=start and end]{red, opacity=0.15};
            \addplot +[mark=none,name path=start,red] coordinates {(0.87, 0) (0.87, 1)};
            \addplot [mark=none,name path=end,red] coordinates {(0.95, 0) (0.95, 1)};
            \tikzfillbetween[of=start and end]{red, opacity=0.15};
            \end{axis}
            % \legend{,pred.,,,,};
            % \end{axis}
        \end{tikzpicture}
        \caption{\textbf{Instrinsic anomaly example}: The system signal is anomalous w.r.t the environment. The dependence structure between environmental varaibles and system variables is broken.}
        \label{fig:intr_anom}
    \end{subfigure}
    
    \caption{Structure of time series corpora (left) and different types of anomalies (middle and right).}
    \label{fig:anomaly_types}
\end{figure*}

Modern applications, appliances, or systems in the physical and virtual world are equipped with sensors that measure and monitor on the one hand the \emph{state of a system} and its sub-components, and on the other hand the \emph{environment} with which the system interacts. These measurements are typically captured by monitoring systems as multi-variate time series data sets. Examples can be found in the Internet-of-Things (IoT) or in the DevOps/AIOps space~\citep{turbines,smartmeter,8814585,aiops,krupitzer2020survey}. Our running example is that of a wind turbine; see Figure~\ref{fig:anomaly_types}. Environmental variables correspond to ambient temperature, wind speed and direction and system variables correspond to rotator speed and internal temperature measurements, among many others. We expect internal temperature measurements to vary naturally with ambient temperature, but a deviation from the functional relationship between ambient and internal temperature measurements could point to a problem of the state of the turbine. It is this change in the functional relationship that is the principal object of interest of this work. Detecting it allows to understand deteriorations of state and health of the system independent of the environment. 

%%!TEX root = ../rewrite.tex

To further motivate our approach, consider a simplified system where \emph{intrinsic normality} means that a single variable $y$ that represents the state of a system depends instantaneously on a single environmental signal $x$ via a mapping $h$, i.e., we have 
$y = h(x) + \varepsilon$, where $\varepsilon$ is a (not further specified) error term. 
We then define an \emph{intrinsic anomaly} as a point where this functional relationship is broken, i.e. $\varepsilon$ exceeds a threshold $\tau$. The ideal signal to detect an intrinsic anomaly is the residual $\delta \coloneqq y - h(x)$. Under normal operation this signal $\delta$ carries no information about the environmental variable $x$ and thresholding the magnitude of this residual signal allows to define intrinsic anomalies. 
 
We contrast intrinsic anomalies with anomalies that we would find if we considered a standard multi-variate anomaly detection problem on environment variables only, which we refer to as \emph{extrinsic anomalies}.
Such problems are widely studied (see ~\citet{blazquezgarcia2020review} for a recent overview). In the wind turbine example, an extrinsic anomaly could be unexpectedly strong wind. Detecting it automatically allows to pitch out the blades if they rotate too fast. However, the pitching out of blades is not an intrinsic anomaly as it constitutes expected behaviour. Figure~\ref{fig:anomaly_types} provides an illustration of both anomaly types. Standard anomaly detection methods~\citep{omnianomaly,liu2008isolation,breunig2000lof} allow to detect extrinsic anomalies, hence we refrain from studying them further here.

The data sets for the intrinsic anomaly detection task differ from standard anomaly detection benchmarks by their division into environment and system variables. 
Similar to standard anomaly detection benchmarks~\citep{bogatinovski2021artificial,Wu_2021}, sparsity and subjectivity of their associated labels for anomalies are practical concerns. 
Sparsity may be due to diverse reasons ranging from practical (e.g., data integration or cost of labelling) to fundamental (internal system failures are exceedingly rare). Subjectivity stems from the fact that failure identification often requires human insight or that alarms come from overly noisy rule-based systems~\citep{bogatinovski2021artificial,Wu_2021}. Based on our practical experience, it is unrealistic to obtain a golden data set since both extrinsic and intrinsic anomalies are domain-specific and subjective. Methods for intrinsic anomaly detection should allow to iteratively improve the data. A common task is checking whether time series similar to a labelled intrinsic anomaly are labelled as such, and if not to potentially (re)label them.

The desiderata for methods for intrinsic anomaly detection are hence as follows. They need to (i) detect intrinsic anomalies and separate them from extrinsic anomalies. While an oversimplification, we will consider the problem of detection extrinsic anomaly detection as solved for the sake of this paper; (ii) sparsity of labels means that we cannot expect a supervised method to work well and instead must rely mostly on unsupervised methods; and (iii) the subjectivity of labels motivates the need for methods to facilitate human interaction with a potentially high-dimensional, complex data set in an interactive and iterative fashion (e.g., by being able to identify similar time series snippets or similar extrinsic/intrinsic anomalies).
Given these desiderata, a natural choice for a method for intrinsic anomaly detection would be based on unsupervised representation learning which 
enables the computation of distances in the embedding space to identify related time series interactively.  

To summarize, our contributions are three-fold. First, we present and formalize a new anomaly detection task in multi-variate time series: intrinsic anomaly detection (Sec.~\ref{sec:problems}). Second, we present a method to tackle intrinsic anomaly detection (Sec.~\ref{sec:method}) which satisfies the afore-mentioned desiderata. In particular, we propose a novel adaptation of unsupervised time series representations~\citep{franceschi2019unsupervised} yielding environment-invariant embeddings.
Third, we provide an initial set of diverse data sets that allows to study this novel problem in a meaningful way: wind turbines as an IoT use case, a semi-synthetic DevOps data set, a swinging pendulum simulation, and a fully synthetic data set based on simple generative models. Our experiments in Sec.~\ref{sec:evaluation} show that our approach based on environment-invariant embeddings performs favorably when compared against standard anomaly detection methods as well as natural baselines.
\section{Intrinsic Anomaly Detection: Problem Statement}\label{sec:problems}

Let $\zts \in \mathbb{R}^{D \times T}$ be a multi-variate, equally spaced time series of length $T$ with $D$ dimensions corresponding to a system/environment combination. 
We denote by $\zts_{t:t+j} = (\zts_{t}, \zts_{t+1}, \ldots, \zts_{t+j})$ a time series snippet of $\zts$. The standard way to define an anomaly at time $t$ for a multi-variate time series is to say that at time $t$ we have an anomaly if $\zts_t$ differs from its expected value, denoted by $\hat{\zts}_t$, by more than a threshold $\tau$~\citep{blazquezgarcia2020review}. A canonical choice in anomaly detection for multi-variate time series is a predictive time series model that, given the history of the multi-variate time series, provides an estimate $\hat{\zts}_t$ via a forecast or a probabilistic generative model. 

In the context of this paper, we have additional structure that we can take advantage of by assuming a decomposition $\zts = (\xts,\yts)$ such that time series $\xts \in \mathbb{R}^{N \times T}$ allows to predict time series $\yts \in \mathbb{R}^{M \times T}$, $D = N + M$, via a function $h$ (modulo an error term $\varepsilon$):
\begin{equation}\label{eq:function_g}
    \yts_t = h(\xts_{0:t}; \yts_{0:t-1}) + \varepsilon \, .
\end{equation}
For all $t$ we have the residual $\delta_t = d(\yts_t, h(\xts_{0:t}; \yts_{0:t-1}))$ for a distance metric $d$. We call $\xts$ the set of \emph{environmental} and $\yts$ the set of \emph{system} time series. 

We define an \emph{extrinsic anomaly} at time $t$ as an anomaly solely in $\xts$, i.e., at time $t$ we have $d(\xts_t, \hat{\xts}_t) > \tau_x$ for a threshold $\tau_x$. At the same time, the functional dependency between $\xts$ and $\yts$ remains unchanged, meaning that the residual signal $\delta_t$ does not exhibit an anomaly: $\delta_t \leq \tau_y$. Conversely, an \emph{intrinsic anomaly} at time $t$ consists of an anomaly in the residual signal $\delta_t$ itself, i.e.,
for a threshold $\tau_y$ we have $d(\yts_t, \hat{\yts}_t) = d(\yts_t, h(\xts_{0:t}, \yts_{0:t-1} )) > \tau_y$. 

We have not further prescribed what $h$ is apart from being a multi-variate function; $h$ may be a  highly non-linear and/or stochastic function that depends on the state of the system as well as (lagged) environmental variables. Depending on the application scenario, $h$ is either given or needs to be obtained. For example, we may have a known physical model describing a system's interaction with the environment or we may assume the need to learn a complex function $h_V \approx h$ from data. In the former case, this relationship between environment and system state will hold also for extreme values in the environment (an extrinsic anomaly), but break whenever the system exhibits internal faults. In the latter case, we may be able to reliably learn parameters $V$ for $h_V$ assuming that intrinsic anomalies are rare. In the data sets that we consider, it is a realistic assumptions that we must learn $h_V$, so we focus on this scenario. 
\section{Representation Learning for Intrinsic Anomaly Detection}
\label{sec:method}

In designing methods for intrinsic anomaly detection, we recall desiderata from the introduction: they must be able to separate intrinsic from extrinsic anomalies, rely on unsupervised learning, and allow to identify similar time series effectively in data quality improvement tasks. We assume that the decomposition into environmental and system time series is given (and does not need to be identified), however we do not assume the functional dependence $h$ to be known.

Our approach relies on two main ingredients: (i) learning effective representations via contrastive learning; and (ii) equipping them to distinguish intrinsic and extrinsic anomalies by forcing representations to be invariant to the environment. We motivate both ingredients separately in Sec.~\ref{sec:contrastive_anomaly_detection} and~\ref{sec:context_invariance} respectively and then present the proposed models. 
 
\subsection{Contrastive Losses and Anomaly Detection}\label{sec:contrastive_anomaly_detection}
Time series representation learning techniques such as~\citep{franceschi2019unsupervised} aim to provide general-purpose representations that are independent of the downstream task. However, there is a particularly close connection to popular approaches in anomaly detection. For example, \citet{ruff18a} propose to learn representations of data points whose enclosing hyper-sphere has minimal size, formally
\begin{equation}\label{eq:one_class}
    \min_W \sum_i (f_W (\zts_{i:i+\ell}) - \cts)^2 + \Omega \,.
\end{equation}
Here, $\zts_{i:i+\ell}$ are contiguous time series snippets of length $\ell$ of $\zts$, $f_W$ is an embedding function $R^{D \times \ell} \rightarrow R^d$ parameterized by $W$, $\cts \in R^d$ is a given center of a hyper-sphere, and $\Omega$ is a regularization term. \citet{ruff18a} argue that under the common and realistic assumption that anomalies are rare, a reasonable embedding function $f_W$ would be learnt such that embeddings of normal time series are close to the cluster center $\cts$ while abnormal ones have a larger distance to $\cts$. The distance between $f_W(\zts_{i:i+\ell})$ and $\cts$ can hence be treated as a metric for the normality of a time series in the embedded space. 

The connection to contrastive learning is then intuitively clear as they construct embeddings equipped with a distance function that fulfills a similar purpose. Instead of choosing a cluster center explicitly,\footnote{\citet{ruff18a} use the average of the initially found embeddings of the training set vectors in the first iteration of the optimization algorithm as cluster centers.} typical contrastive losses flexibly choose points that are (heuristically) close to each other (e.g., close in time for time series). We formalize this as
\begin{equation}\label{eq:positive_examples}
    \min_W \sum_i \sum_j d(f_W (\zts_{i:i+\ell}), f_W(\zts_{j:j+k})) + \Omega \, ,
\end{equation}
where $d$ is a distance function in the embedding space (e.g., $\ell_2$ as in~\shorteqref{eq:one_class}). Typically, the cluster centers $\zts_{j:j+k}$ are referred to as \emph{reference} examples and $\zts_{i:i+\ell}$ are \emph{positive} examples with $i, \ell$ randomly chosen such that $\zts_{i:i+\ell} \subseteq \zts_{j:j+k}$.

In order to guarantee the second property of~\shorteqref{eq:one_class} -- that abnormal points are far from each other -- many contrastive losses explicitly require that points that are (heuristically) different are further away in the embedding space. These \emph{negative} examples should be far apart in terms of the cluster centers, leading to the desired property that individual hyper-spheres have small volume (normal values are concentrated) and hyper-spheres among themselves are subject to a repelling effect (summarized by $\Omega$ in~\shorteqref{eq:positive_examples}).

\subsection{Representations for Intrinsic Anomaly Detection}\label{sec:context_invariance}
While contrastive losses allow for standard anomaly detection in multi-variate time series (see e.g.,~\citet{blazquezgarcia2020review} for an overview), they do not allow fine-grained distinctions into intrinsic and extrinsic anomalies. Following the arguments above, we expect extrinsic and intrinsic anomalies to correspond to outliers in the embedding space because they are rare. It is unclear however why an embedding of an extrinsic anomaly, should necessarily be distant from an embedding of an intrinsic anomaly. To guarantee that we can separate extrinsic and intrinsic anomalies in the embedding space, it is natural to consider the inclusion of an inductive bias, namely an invariance of the representations towards the environment, to ensure this separation. First, note that embeddings that are invariant to changes in the environment do not carry information about the environment, analogous to the ideal residual signal in Sec.~\ref{sec:problems}. For an environment-invariant embedding, intrinsic normality corresponds to proximity in the embedding space (because this is the frequent case), while intrinsic anomalies violate the standard invariance and would hence be more distant in the embedding space. Second, we can modify contrastive losses explicitly to encode that a broken functional relationship between environment and system variables correspond to greater distance in the embedding space. 

\paragraph{Adversarial learning for Environment-Invariance}
For obtaining representations of $\zts$ that are invariant to the environment $\xts$, a popular choice is to apply an adversarial learning approach~\cite{xie2018controllable,long2017conditional} which we adapt to our unsupervised setting. We train the predictor $g_U$ in an adverserial manner jointly with the embedding model: The predictor's parameters are trained to maximize the likelihood of $\xts$ given embeddings  $f_W(\zts)$; the weights of the embedding model $f_W$ are trained to maximize the error of the predictor. This encourages the embeddings to become invariant to $\xts$. Note that for the prediction task, we use a multi-class classification approach by discretizing $\xts$ appropriately. Hence, we compute the negative log probability of the correct class as our loss, i.e.,
\begin{equation}\label{eq:loglikelihood}
    \mathcal{L}_\text{adv}( g_U(f_W (\zts_{i:j})), \xts_{i:j}) = - \log \frac{1}{g_U(f_W( \zts_{i:j}))_{\xts_{i:j}}}.
\end{equation}
For $g_U$ we use a neural network with a softmax function as the final layer. The full loss is given by
\begin{equation}\label{eq:env_invariance}
\mathcal{L}_\text{adv}(W, U) = \sum_{i,j}\mathcal{L}_\text{adv}( g_U(f_W (\zts_{i:j})), \xts_{i:j}) \,.
\end{equation}

\paragraph{Negative Examples in Contrastive Losses} For embeddings of $\zts$, we can equip~\shorteqref{eq:positive_examples} with negative examples that explicitly encode when the dependency structure is broken. We construct these negative examples as system and environment variables originating from different time snippets. We denote $\zts^{\pos} = (\xts^{\pos}, \yts^{\pos}) = (\xts_{i:i+\ell},\yts_{i:i+\ell})$ as a positive and $\zts^{\ctx} = (\xts^{\ctx}, \yts^{\ctx}) = (\xts_{j:j+k},\yts_{j:j+k})$ as a reference time series snippet where $i,j,k,\ell$ are chosen (randomly) such that positive examples are contained in the reference time series. In the construction of negative examples, we explicitly break the dependency structure between $\xts$ and $\yts$ in defining $\zts^{\nega} = (\xts^{\nega}, \yts^{\pos}) =  (\xts_{r:r+\ell}, \yts_{i:i+\ell})$ where $r$ is (randomly) chosen such that $\xts_{\nega} \cap \xts_{\pos} \approx \emptyset$. Figure~\ref{fig:sample_selection} provides an illustration. The contrastive loss uses these three sample types such that time series with default dependence structure should be embedded close to each other while time series with varying dependence structures should be distant from each other. For an embedding network $f_W$ with parameters $W$, the loss function takes the following form:
\begin{align}\label{eq:contrastive}
    \begin{split}
    \mathcal{L}_\text{contr}(W) = & \sum_{\pos} \sum_{\ctx} d(f_{W}( \zts^{\pos} ), f_{W}( \zts^{\ctx}) )\ - \sum_{\nega} \sum_{\ctx} d(f_{W}( \zts^{\nega} ), f_{W}( \zts^{\ctx} ) )\;.
    \end{split}
\end{align}
We aim to minimize~\shorteqref{eq:contrastive}. Note that~\shorteqref{eq:contrastive} is amenable to batch optimization and we can further construct negative examples such that negative environment variables are chosen from a different series in the batch.

\subsection{Model Overview}
\label{sec:embed}

Our approach consists of two main components: a \emph{predictor network} $g_U$ for environment invariance, and an \emph{embedding network} $f_W$ which learns embeddings using contrastive losses. 
Figure~\ref{fig:architecture} depicts the main components of our approach. For $g_U$, we choose a single-layer feed-forward neural network to place more onus on $f_W$ to become invariant to the environment. For discretization of $\xts$, we choose standard quantile-based binning with 20 buckets. 

The basic building block of our embedding network architecture consists of stacked temporal dilated causal convolutions~\citep{tcn} following~\citep{franceschi2019unsupervised}. We choose such a temporal convolutional neural network~(TCN) for its speed of computation and its general robustness to facilitate experimentation.

\begin{figure*}[ht]
    \centering
	
%	\node[] (un) at (-7.5,0) {Units};
%    \node[] (sen) at (-7.5,-2) {Signals};
%    \node[] (stat) at (-7.5,-4) {Statistics};
%    
%    \node[draw] (u1) at (0,0) {\includegraphics[width=0.05\textwidth]{figs/turbine.pdf} Turbine 1};

\begin{subfigure}[t]{0.3\textwidth}
        \centering
        \scalebox{0.7}{
        \begin{tikzpicture}
    
    \path[fill=white] (1,-1) rectangle ++(2.5,2.5);
    
	\draw[draw, fill=Orange!10] (0,0) rectangle ++(5.2,0.6);
    \draw[draw, fill=Cyan!10] (0,0.6) rectangle ++(5.2,0.6);
    \draw[draw, fill=Orange!10] (0,-1.35) rectangle ++(5.2,0.6);
    \draw[draw, fill=Cyan!10] (0,-0.75) rectangle ++(5.2,0.6);
    
    \node[Cyan!66!black] (ts1) at (5.75,0.85) {env};
    \node[Orange!66!black] (ts2) at (5.75,0.25) {sys};
    \node[] (zi) at (-0.5,0.6) {$\zts_i$};
    % \draw [decorate,decoration={brace,amplitude=10pt},xshift=-4pt,yshift=0pt] (0.5,0.5) -- (0.5,5.0) node [black,midway,xshift=-0.6cm]
    % \node[Orange!66!black] (ts2) at (-2,0.5) {multivariate time series};
    
    \node[Cyan!66!black] (ts1) at (5.75,-0.5) {env};
    \node[Orange!66!black] (ts2) at (5.75,-1.1) {sys};
    \node[] (zj) at (-0.5,-0.85) {$\zts_j$};
    % \node[Orange!66!black] (ts2) at (-2,-0.75) {multivariate time series};
    
	\node[draw, minimum width=70pt, minimum height=17pt,fill=refcolor] (c) at (1.6,0.9) {\hspace*{50pt}$\xts{^\ctx}$};
	\node[draw, minimum width=70pt, minimum height=17pt, fill=refcolor] (d) at (1.6,0.3) {\hspace*{49.85pt}$\yts{^\ctx}$};
	\definecolor{greenlocal}{RGB}{21,152,76}
	\definecolor{redlocal}{RGB}{214,38,49}
	\node[draw, minimum width=30pt, minimum height=17pt,fill=poscolor] (a) at (1.25,0.9) {\hspace*{5pt}$\xts{^\pos}$\hspace*{5pt}};
	\node[draw, minimum width=30pt, minimum height=17pt, inner sep=3.5pt, fill=white, left color=greenlocal, right color=redlocal] (b) at (1.25,0.3) {\hspace*{4.75pt}$\yts^{\pos}$\hspace*{4.75pt}};
	\node[draw, minimum width=30pt, minimum height=17pt,fill=negcolor] (x) at (4.3,-0.45) {\hspace*{5pt}$\xts^{\nega}$\hspace*{5pt}};

	\node[draw, minimum width=30pt, minimum height=15pt,fill=refcolor] () at (0.5,3.5) {$\xts^{\ctx}$};
	\node[draw, minimum width=30pt, minimum height=15pt, fill=refcolor] (s) at (0.5,2.95) {$\yts^{\ctx}$};
	
	\node[black,align=center] () at (0.5,2) {reference\\ sample};
	
	\node[draw, minimum width=30pt, minimum height=15pt,fill=poscolor] () at (2.75,3.5) {$\xts^{\pos}$};
	\node[draw, minimum width=30pt, minimum height=15pt, fill=poscolor] () at (2.75,2.96) {$\yts^{\pos}$};
	\node[black,align=center] () at (2.75,2) {positive\\ sample};
	
	\node[draw, minimum width=30pt, minimum height=15pt,fill=negcolor] () at (5,3.5) {$\xts^{\nega}$};
	\node[draw, minimum width=30pt, minimum height=15pt, fill=negcolor] () at (5,2.96) {$\yts^{\pos}$};
	\node[black,align=center] () at (5,2) {negative\\ sample};
	
	\end{tikzpicture}
	}

	\definecolor{greenlocal}{RGB}{21,152,76}
	\definecolor{redlocal}{RGB}{214,38,49}
	\caption{\textbf{\textcolor{Cyan}{Reference}, \textcolor{greenlocal}{positive}, and \textcolor{redlocal}{negative} sample selection overview}. While reference and positive sample selection corresponds to~\citet{franceschi2019unsupervised}, negative samples are selected such that the dependence structure between environment and system signals is explicitly broken. \label{fig:sample_selection}
	%This is achieved by choosing an environment example outside the context range, $k$, while choosing a positive within the context range, $b = l$. The color gradient in $b = l$ indicates that this particular time series frame is part of both a positive and a negative sample.
	}
    \end{subfigure}
    \hspace{10pt}
    \begin{subfigure}[t]{0.65\textwidth}
        \centering
        \scalebox{0.8}{
        
        \begin{tikzpicture}
	\draw[draw, lightgray, fill=white] (-2.5,1.25) rectangle ++(7.5,2.5);
	\draw[draw, lightgray, fill=white] (5.5,1.25) rectangle ++(5.5,2.5);
    
    \node[text width=2cm,align=center] (emb) at (-1.5,5) {Embedding space with contrastive losses};
    \node[text width=2cm,align=center] (ctr) at (-1.1,2.5) {Embedding component};
    \node[text width=2cm,align=center] (da) at (6.7,2.5) {Predictive component with environment adversary};
    %\node[] (enc) at (-1.5,3.25) {Encoder};

	\draw[draw, fill=white] (0,3.95) rectangle ++(11,2);

	\node [trapezium, trapezium angle=60, draw] (enc1) at (1,3.25) {$f_W$};
	\node [trapezium, trapezium angle=60, draw] (enc2) at (2.5,3.25) {$f_W$};
	\node [trapezium, trapezium angle=60, draw] (enc3) at (4,3.25) {$f_W$};
	
	\node [trapezium, trapezium angle=-60, draw] (pred1) at (8.5,3) {$g_U$};
	\node [trapezium, trapezium angle=-60, draw] (pred2) at (10,3) {$g_U$};
	%\node [trapezium, trapezium angle=60, draw] (enc3) at (4,3.25) {$f_W$};
	
% 	\node[draw, inner sep=0.38cm, fill=refcolor] (ref) at (3.5,0.5) {\hspace*{100pt}ref};

	\definecolor{greenlocal}{RGB}{21,152,76}
	\definecolor{redlocal}{RGB}{214,38,49}

% 	\node[draw, inner sep=0.38cm, fill=poscolor] (pos) at (3.75,0.5) {\hspace*{10pt}pos\hspace*{10pt}};
% 	\node[draw, minimum width=50pt, minimum height=14pt, fill=negcolor] (y) at (8.5,-1.0) {\hspace*{10pt}$y$\hspace*{10pt}};
	
	\node[draw, minimum width=30pt, minimum height=15pt,fill=refcolor] (c_1) at (1,2.25) {$\xts^{\ctx}$};
	\node[draw, minimum width=30pt, minimum height=15pt, fill=refcolor] (d_1) at (1,1.72) {$\yts^{\ctx}$};
	
	\node[draw, minimum width=30pt, minimum height=15pt,fill=poscolor] (a_1) at (2.5,2.25) {$\xts^{\pos}$};
	\node[draw, minimum width=30pt, minimum height=15pt, fill=poscolor] (b_1) at (2.5,1.72) {$\yts^{\pos}$};
	
	\node[draw, minimum width=30pt, minimum height=15pt,fill=negcolor] (x_1) at (4,2.25) {$\xts^{\nega}$};
	\node[draw, minimum width=30pt, minimum height=15pt, fill=negcolor] (b_2) at (4,1.72) {$\yts^{\pos}$};
	
	\node[draw, minimum width=30pt, minimum height=15pt,fill=negcolor] (x_2) at (8.5,2) {$\xts^{\nega}$};
	\node[draw, minimum width=30pt, minimum height=15pt, fill=poscolor] (a_2) at (10,2) {$\xts^{\pos}$};
	
	\node[negcolor, circle, fill=negcolor, minimum size=0.05cm] (x_neg) at (9.5,5.5) {};% {x};
	\node[refcolor, circle, fill=refcolor, minimum size=0.05cm] (x_ref) at (1.5,5.5) {};% {x};
	\node[poscolor, circle, fill=poscolor, minimum size=0.05cm] (x_pos) at (2.5,4.5) {};% {x};
    
    \draw [->, refcolor, thick] (c_1) to (enc1); %, decorate, decoration=snake
    \draw [->, poscolor, thick] (a_1) to (enc2);
    % \draw [->, refcolor, thick] (c_1) to (enc1);
    % \draw [->, poscolor, thick] (a_1) to (enc2);
    \draw [->, negcolor, thick] (x_1) to (enc3);
    \draw [->, negcolor, thick] (enc3) to [out=90,in=270] (x_neg);
    \draw [->, poscolor, thick] (enc2) to (x_pos);
    \draw [->, refcolor, thick] (enc1) to [out=90,in=180] (x_ref);
    
    \draw [->, poscolor, thick, dotted] (x_pos) to [out=0,in=90] (pred1);
    \draw [->, negcolor, thick, dotted] (x_pos) to [out=0,in=90] (pred2);
    \draw [->, poscolor, thick, dotted] (pred1) to (x_2);
    \draw [->, negcolor, thick, dotted] (pred2) to (a_2);
    
    \draw [<->, black, thick, dashed] (x_neg) to (x_ref);
    \draw [>-<, black, thick, dashed] (x_pos) to (x_ref);
	
	\end{tikzpicture}
	}
	\caption{\textbf{Architecture overview with contrastive and adversarial learning for environment invariance}. At its core, we employ a contrastive learning component that pushes reference and positive samples closer to each other in the embedding space while trying to push context and negative samples apart. 
	%Note that reference and positive samples are augmented at random, which is denoted by the squiggly arrows. 
	We further have an adversarial component to obtain environment invariance that, given a positive embedded sample, tries to reconstruct the context of the positive sample as badly as possible (red dotted arrows) while reconstructing the context of the negative sample as well as possible (green dotted arrows).
	\label{fig:architecture}
	}
    \end{subfigure}
    %\vspace{-10pt}
	\caption{Overview of the sample selection process for contrastive learning and architecture diagram.}
	\label{fig:da_overview}
\end{figure*}

The final loss is a combination of a contrastive loss with a regularizer to obtain environment invariance, which is summarized as follows:
 \begin{equation}\label{eq:combined-loss}
\mathcal{L}(W,U) = \mathcal{L}_\text{contr}(W) - \lambda \mathcal{L}_\text{adv}(W, U)\, ,
 \end{equation}
 where $\mathcal{L}_\text{contr}(W)$ is the contrastive loss~\shorteqref{eq:contrastive}, $\mathcal{L}_\text{adv}(W, U)$ is the environment-invariance loss~\shorteqref{eq:env_invariance}, and $\lambda$ is a hyperparameter that governs the trade-off between them.  We seek a saddle point $(\hat{W}, \hat{U})$ such that $\hat{W} = \argmin_{W} \mathcal{L}(W, \hat{U})$ and $\hat{U} = \argmin_U \mathcal{L}(\hat{W}, U)$. 
We use a gradient-reversal layer between $f$ and $g$ to identify the saddle point in~\shorteqref{eq:combined-loss} as proposed in~\citep{ganin2016domain}. 
\citet{moyer18} provide further theoretical and practical justifications for it.
\section{Experiments}
\label{sec:evaluation}

In our experiments, we empirically evaluate our approach both quantitatively and qualitatively. Public time series data sets for (standard) anomaly detection\footnote{\citet{Wu_2021} convincingly argue that many of these datasets should be abandoned.} are not suitable for intrinsic anomaly detection, so we do not consider them. Instead, we evaluate on four different data sets, each of which allows for a separation into environmental and system signals. We discuss these data sets first, then describe our evaluation approach and the models under consideration, and finally present our results. 

\begin{table}
 \centering
  \vspace{5pt}
  \scriptsize
  \setlength{\tabcolsep}{7pt}
    \caption{Data set summary statistics. Each dataset consists of multiple distinct sub-systems. For each system, we have a number of time series snippets available with length $T$. $N$ and $M$ denote the number of environmental and system variables, respectively.}
  \begin{tabular}{cccccccc}
    \toprule
    Dataset & \# Systems & \# Series & $T$ & $N$ & $M$  \\
    \midrule
    \synthetic & 1 system & 360 & 1440 & 2 & 2 \\
    \pendulum & 1 pendulum & 300 & 144 & 1 & 2  \\
    \turbine & 4 turbines & 360 & 144 & 5 & 4 \\
    \devops & 1 app & 128 & 128 & 1 & 3 \\
    \bottomrule
  \end{tabular}
  \label{tab:datasets}
\end{table}

\subsection{Datasets}
The selection of benchmark datasets aims to balance physical and virtual systems as well as synthetic and real-world data.  
Note that synthetic data exhibits perfect labels, while real-world data typically does not.  
While unfortunate, we believe that this subjectivity and noisiness must be embraced as fundamental in the task. Hence, we place additional emphasis on qualitative experiments. Appendix~\ref{app:data_details} contains further details and illustrations on the data sets and Table~\ref{tab:datasets} captures details of the datasets under consideration. 

\paragraph{Synthetic} We generate a synthetic dataset based on a simple sinusoidal generative model. We generate two environmental signals, as well as two system signals. We inject two types of anomalies into the data: (i) extrinsic anomalies, i.e., anomalies in the environmental which are also instantly reflected in the system variables and (ii) intrinsic anomalies only in the system variables. 

\paragraph{Pendulum} We consider the case of a swinging pendulum with added control signals, where we control the dampening of the acceleration from the outside as an environmental signal (towards which we want to be invariant). As intrinsic anomalies, we consider those where we inject an anomaly as a change in the length of the chord, which we capture as part of the system signal. Our aim with this data set is to understand how well our models handle cases where the dependency structure between system and environment is significantly more complex (i.e. time-shifted and non-linear) than in the \synthetic\ data set. 

\paragraph{DevOps} is a new, semi-synthetic data set\footnote{We will open-source both the raw data as well as the set-up to produce the data.} 
which resembles data sets behind corporate firewalls. It is based on a cloud-based microservice demo app,\footnote{\url{https://github.com/microservices-demo/microservices-demo}} commonly used in a DevOps context~\citep{sockshop}. The environment for the application are user interactions with the app, approximated by the network outbound traffic of an auxiliary application that induces synthetic load/user behaviour on the app. The system signals consist of metrics such as CPU consumption, memory or disk consumption of the app. 
We inject anomalies both in the user behavior which lead to extrinsic anomalies  
and in the internal state to obtain intrinsic anomalies.
Note that although we have almost total control of the application, the labelling of anomalies is still imperfect, thereby adding further to the complications of public anomaly detection benchmarks~\citep{Wu_2021}. 

\paragraph{Wind Turbine} is a public wind turbine data set by Energias de Portugal.\footnote{\url{https://opendata.edp.com/pages/homepage/}} The time series panel can be separated into environmental signals (e.g., wind speed/direction, ambient temperature) and system signals (e.g., rotational speeds of generators, temperature on different components, power output). 
Note that this data set contains only few (43) labelled anomalies and visual inspections of the data reveals inconsistencies with these labels, see e.g., Figure~\ref{fig:incorrectly_labelled}. 
Hence, purely quantitative evaluations cannot be taken at face value. Operating under this limitation, we illustrate the versatility of our approach qualitatively in Sec.~\ref{sec:qualitative}.

\subsection{Model and Evaluation Details}
For a comprehensive empirical comparison, we consider the following families of approaches: (i) approaches based on deep embeddings to varying degree of complexity, the most closely related to our approach; (ii) approaches based on classical time series embeddings; and (iii) standard anomaly detection methods on time series. We consider these approaches on both the original time series and on the residual signal. We further note that we only perform hyperparameter tuning on the \synthetic\ data set as we do not have enough labels available on other datasets. Appendix~\ref{app:more_exp_details} provides more fine-grained details on particular hyperparameter choices for the models described next.

\paragraph{Deep Embeddings} We consider the following models in our comparison which we list in increasing order of complexity. We can choose to ignore the functional dependency $h$ and learn representations of $\zts$ with a TCN equipped with contrastive losses.  
We denote this approach as \basicemb. Next, we propose a two-step approach for environment-invariance by explicitly approximating the functional dependence $h$ between $\xts_t$ and $\yts_t$ using a  neural network $h_V$ first, and  
then learning embeddings of the \emph{residual} signal $\hat{\delta} := h_V(\xts) - \yts$ in a second step. We refer to this approach as \twostepmodelregr. Finally, we use the approach outlined in Sec.~\ref{sec:method} which we call \domainadv. We choose $\lambda=10^{-3}$ as determined by an ablation study (see Appendix~\ref{app:more_exp_details}). 

\paragraph{Classical Embeddings}
We compare against two instances of~\citep{lubba2019catch22}: \catchtwotwo\ and \residualcatchtwotwo\ which compute a feature vector per original and residual time series respectively similar to \twostepmodelregr. 

\paragraph{Anomaly Detection Methods} We consider a simple thresholding mechanism, \residualthresh, which thresholds the residual signal $\hat{\delta}$ that is also obtained in \twostepmodelregr; \ominanomaly~\citep{omnianomaly} as a popular example for deep anomaly detection; and as examples for classical approaches: Isolation Forests~\citep{liu2008isolation} \isolationforest, Local Outlier Factor~\citep{breunig2000lof} \localoutlierfactor{}, and One-class SVM~\citep{scholkopf1999support} \ocsvm.

\paragraph{Evaluation Details} For each dataset, we break the original multi-variate time series into time series snippets (length described in Table~\ref{tab:datasets}) and classify each snippet as abnormal if it contains an intrinsic anomaly. Although the length of an intrinsic anomaly is variable, intrinsic anomalies appear as intervals in our data sets and not as isolated single-point anomalies. While embedding-based approaches can handle variable-length input series, classic embeddings and traditional anomaly detection approaches provide an anomaly score for each time step. To compare these methods on an equal footing, we aggregate anomaly scores from individual time steps to a single score for a full snippet. We report mean results over 5 random seeds.

\begin{table*}[!t]
    \centering
    \scriptsize
    \setlength{\tabcolsep}{4pt}
    \caption{AUROC results on the intrinsic anomaly detection task.}
    \begin{tabular}{ccccccc}
        \toprule
        Category & Signal Type & Detection Approach & \synthetic & \pendulum & \devops & \turbine \\
        \midrule
        \multirow{3}{*}{Deep Embeddings} & \multirow{2}{*}{Original} & \domainadv & \textbf{0.999} ($\pm$ 0.002) & \textbf{0.980} ($\pm$ 0.002) & 0.587 ($\pm$ 0.007) & 0.756 ($\pm$ 0.022) \\
        & & \basicemb & 0.512 ($\pm$ 0.022) & 0.969 ($\pm$ 0.013) & 0.535 ($\pm$ 0.041) & 0.632 ($\pm$ 0.015) \\
        \cdashlinelr{2-7}
        & Residual & \twostepmodelregr & \textbf{1.000} ($\pm$ 0.000) & 0.951 ($\pm$ 0.015) & 0.532 ($\pm$ 0.036) & 0.725 ($\pm$ 0.018) \\
        \midrule
        \multirow{2}{*}{Classic Embeddings} & Original & \catchtwotwo & 0.494 ($\pm$ 0.008) & 0.904 ($\pm$ 0.000) & 0.573 ($\pm$ 0.000) &  0.512 ($\pm$ 0.000)\\
        \cdashlinelr{2-7}
        & Residual & \residualcatchtwotwo & \textbf{1.000}  ($\pm$ 0.000) & 0.891 ($\pm$ 0.000) & 0.577 ($\pm$ 0.000) & 0.680 ($\pm$ 0.000) \\
        \midrule
        \multirow{9}{*}{Anomaly Detection} & \multirow{4}{*}{Original} & \localoutlierfactor & 0.557 ($\pm$ 0.016) & 0.541 ($\pm$ 0.042) & 0.518 ($\pm$ 0.033) & 0.578 ($\pm$ 0.026) \\
        & & \isolationforest & 0.500 ($\pm$ 0.014) & 0.502 ($\pm$ 0.024) & 0.489 ($\pm$ 0.013) & 0.512 ($\pm$ 0.018) \\
        & & \ocsvm & 0.499 ($\pm$ 0.022) & 0.500 ($\pm$ 0.013) & 0.569 ($\pm$ 0.025) & 0.521 ($\pm$ 0.017) \\
        & & \ominanomaly & 0.788 ($\pm$ 0.108) & 0.642 ($\pm$ 0.082) & 0.513 ($\pm$ 0.023) & 0.561 ($\pm$ 0.026) \\
        \cdashlinelr{2-7}
        & \multirow{5}{*}{Residual} & \localoutlierfactor & 0.537 ($\pm$ 0.046) & 0.467 ($\pm$ 0.030) & 0.573 ($\pm$ 0.070) & 0.679 ($\pm$ 0.045) \\
        & & \isolationforest & 0.513 ($\pm$ 0.020) & 0.525 ($\pm$ 0.015) & 0.507 ($\pm$ 0.008) & 0.534 ($\pm$ 0.012) \\
        & & \ocsvm & 0.546 ($\pm$ 0.012) & 0.534 ($\pm$ 0.009) & 0.571 ($\pm$ 0.013) & 0.533 ($\pm$ 0.010) \\
        & & \ominanomaly & 0.801 ($\pm$ 0.075) & 0.717 ($\pm$ 0.113) & 0.435 ($\pm$ 0.028) & 0.492 ($\pm$ 0.067) \\
        & & \residualthresh & \textbf{1.000}  ($\pm$ 0.011) &  0.510 ($\pm$ 0.000) & \textbf{0.619} ($\pm$ 0.000) & \textbf{0.845} ($\pm$ 0.032) \\
        \bottomrule
    \end{tabular}
    \label{tab:results_auroc}
\end{table*}

\begin{table}[!t]
  \centering
  \setlength{\tabcolsep}{5pt}
  \caption{The $\ell_2$ distance gap between embeddings of correct and incorrect classes on the anomaly detection task. Environment-invariant embeddings induce the largest gap.}
  \begin{tabular}{ccccc}
    \toprule
    & \basicemb & \twostepmodelregr & \catchtwotwo & \domainadv \\
    \midrule
    \synthetic & 0.051 ($\pm$ 0.041) & 0.641 ($\pm$ 0.314) & 0.008 ($\pm$ 0.01) & \textbf{1.944} ($\pm$ 0.039) \\
    \pendulum & \textbf{0.446} ($\pm$ 0.059) & \textbf{0.433} ($\pm$ 0.074)  & 0.019 ($\pm$ 0.00)  & \textbf{0.486} ($\pm$ 0.178)\\
    \devops & 0.014 ($\pm$ 0.006) & \textbf{0.035} ($\pm$ 0.008) & 0.013 ($\pm$ 0.00) & \textbf{0.042} ($\pm$ 0.007)  \\
    \turbine & 0.075 ($\pm$ 0.007) & \textbf{0.319} ($\pm$ 0.103)  & 0.004 ($\pm$ 0.00) & \textbf{0.450} ($\pm$ 0.085)\\
    \bottomrule
  \end{tabular}
  \label{tab:results_distance}
\end{table}

\begin{table}[!t]
    \centering
    \setlength{\tabcolsep}{3.5pt}
    \caption{F-1 scores for the anomaly type classification task. While primarily designed to capture intrinsic anomalies, the disentanglement yielded by environment-invariant embeddings allows for the identification of extrinsic anomalies.}
    \begin{tabular}{cccccc}
    \toprule
         & \basicemb & \twostepmodelregr & \catchtwotwo & \residualcatchtwotwo & \domainadv \\
        \midrule
        2-class  & 0.85 ($\pm$ 0.04) & \textbf{0.98} ($\pm$ 0.01) & 0.85 ($\pm$ 0.02) & \textbf{0.99} ($\pm$ 0.01) & \textbf{0.99} ($\pm$ 0.01) \\
        3-class  & 0.74 ($\pm$ 0.03) & 0.59 ($\pm$ 0.08) & 0.71 ($\pm$ 0.03) & 0.64 ($\pm$ 0.06) & \textbf{0.89} ($\pm$ 0.03) \\
    \bottomrule
    \end{tabular}
     \label{tab:results_disentanglement}
\end{table}

\input{tikzplots/turb_nn}

\subsection{Quantitative Results}

We study both the raw intrinsic anomaly detection performance as well as the disentanglement of extrinsic and intrinsic anomalies in the learned latent spaces.

\paragraph{Intrinsic Anomaly Detection Performance} Acknowledging recent concern on the validity of time series anomaly benchmarks~\citep{kim2021towards}, our main quantitative results report AUROC scores on the intrinsic anomaly detection task. For the \texttt{ResThresh} method, we compute the AUROC score based on the maximum residual value over the full residual series. For the embedding-based approaches, we use a $k$-nearest-neighbor classifier ($k=5$) in the embedding space to determine a discrete anomaly label.

Table~\ref{tab:results_auroc} summarizes our main quantitative results. First, we note that the difficulty of intrinsic anomaly detection differs widely with the data sets. For example, intrinsic anomaly detection seems relatively easier on the \turbine\ data set compared with the \devops\ data set despite the latter being semi-synthetically generated with controlled labels. Second, we note that \domainadv\ leads to overall superior results when comparing the embedding based approaches (the first three rows in Table~\ref{tab:results_auroc}) and is overall competitive results in many cases, but not always. For the \devops\ and \turbine\ data sets, \residualthresh\ is overall best by a margin. 
The explanation here is that the simple instantaneous mapping of \residualthresh\ is a good inductive bias for these data sets -- however, this is in particular not the case for \pendulum. 
For the \pendulum\ data set, it is worthwhile to note that \domainadv{} delivers an overall superior approach despite the complex non-linear interaction between the environmental and system signals and only a linear environment-predictor component. We further note that our approach is superior to the other anomaly detection methods consistently and independently whether we apply the anomaly detection methods directly on the time series or when taking the residual.

\paragraph{Embedding Space Distances} For methods that produce a mapping onto an embedding space, we can further determine whether any particular (embedded) time series is closer to the mean embedding of the respective correct/incorrect class. Concretely, we can compute the gap between the distance to the correct class for the respective series and the distance to the incorrect class and average this quantity over all series in the test set. Results are presented in Table~\ref{tab:results_distance}. This complements the AUROC results in Table~\ref{tab:results_auroc} by providing an indication of the degree of entanglement in the embedding space and the resulting nearest-neighborhood structure. Overall, we observe that environment-invariant embeddings produce the largest gap. 

\paragraph{Identifiability of Extrinsic vs Intrinsic Anomalies} We further provide results on the identifiability of both extrinsic anomalies and intrinsic anomalies. For the Synthetic dataset, we create two distinct labelings: (i) A \textit{2-class} labeling where both normal series snippets and extrinsic-anomalous series are assigned the label 0 and intrinsic anomalies only are assigned the label 1; and (ii) a \textit{3-class} labeling where normal snippets are labeled as 0, extrinsic anomalies are labeled as 1, and intrinsic anomalies are labeled as 2. Performance on the first labeling indicates how well each method is capable of separating intrinsic anomalies from all encounterable scenarios (1-vs-all setting) while performance on the second labeling indicates the level of disentanglement of individual anomaly types in the embedding space. Note that our evaluation in Table~\ref{tab:results_auroc} relies on 2-class labeling.

We report the weighted F-1 scores for both labeling strategies as part of Table~\ref{tab:results_disentanglement}. We observe that environment-invariant embeddings provide both (i) strong 1-vs-all performance (2-class) and at the same time still (ii) enable superior identification of normal and extrinsic-anomalous series (3-class). We also find that residual-based approaches (i.e. \twostepmodelregr\ and \residualcatchtwotwo) provide strong 1-vs-all performance but they do so at the expense of extrinsic anomaly identification. These findings suggests that environment-invariant embeddings enable us to learn more complex non-linear dependence structures and patterns that go beyond simple residual signals.

\subsection{Qualitative Results: A Case Study On Turbines}\label{sec:qualitative}
The embeddings induce a nearest-neighbor structure which allows us to navigate the time series corpus via distances. This can be helpful in exploring the data set and uncovering data or label issues. In Figure~\ref{fig:nn_turbine}, we show this in qualitative results for environment-invariant embeddings on the Turbine dataset. The first two columns show sanity checks: reference time series (top row) labelled as normal or abnormal have as their nearest neighbors (same column as the reference time series) normal and abnormal time series, respectively. Despite the multi-variate nature of the data, visual inspection confirms the plausibility of the nearest neighbors. The last column in Figure~\ref{fig:nn_turbine} shows an example where a reference time series is labelled as normal, but its nearest neighbors consist of time series that are labeled as abnormal. This could point to a labelling issue, i.e., it may make sense to re-label the reference time series in column (c) as abnormal for the abnormal stretches 0-40 and 100-130.
\section{Related work}

Anomaly detection in time series has a rich body of literature, see e.g.,~\citet{omnianomaly,carmona2021neural,ren19,ayed2020anomaly,Wu_2021,ya19,guha16,blazquezgarcia2020review} for a recent overview. While different variants of anomalies are the subject of discussion, we are not aware of a prior discussion of intrinsic anomalies (see e.g.,~\citet{nature_anomalies_21} for a comprehensive overview of anomaly types). Intrinsic anomaly detection is a subset of the more general class of contextual anomalies where \emph{context} is more broadly and fuzzily defined ~\citep{han2012mining}. The closest relative are conditional anomalies~\citep{4138201} defined for tabular data, however they ignore the interplay between extrinsic and intrinsic anomalies. We are not aware of other publicly available multi-variate time series data sets that lend themselves to targeted study of intrinsic anomaly detection. Existing detection methods can be adopted to the intrinsic anomaly detection problem by relying on the residual signal, however our experiments show that custom-built methods are superior. A possibility for future work is the refinement of our problem statement with a causal treatment, see~\citet{janzing2019causal} for an example outside the time series context as well as discovery of the causal dependency structure~\citep{haufe2009sparse,app10062166,qiu2012granger}. 

Similar to anomaly detection for time series, representation learning for time series has a rich body of literature, e.g.,~\citet{franceschi2019unsupervised,transformerReps,lubba2019catch22,tsfresh}. The connection between anomaly detection and contrastive losses is not new (see e.g.,~\citet{NEURIPS2020_8965f766,sohn2021learning} for recent examples), but it is under-explored in the time series domain. 
While we extend~\citet{franceschi2019unsupervised} for practical reasons such as speed of experimentation and general robustness, we remark our approach carries over also to other architectures, such as~\citet{transformerReps}. 

Arguably, representation learning is less prominently discussed in time series compared to other disciplines such as computer vision or natural language processing~\citep{chen2020simple,oord2019representation,he2020momentum,fang2020cert,jaiswal2021survey}. While these fields consider learning invariant representations a central question, see e.g.,~\citet{ganin2016domain,moyer18,akash2021learning,achille18}, this has not been the case for the time series domain to the best of our knowledge. Correspondingly, work on domain adaptation is in its infancy~\citep{jin2021domain}. Our approach attempts to change this and relies heavily on \citet{xie2018controllable} by considering invariance with respect to variables included in the data. 
\section{Conclusion}
\label{sec:conclusions}

We present a novel variation of the anomaly detection task in multi-variate time series where an intrinsic anomaly manifests itself as a departure from the typical functional relationship between an environment and a system. Our main method for addressing intrinsic anomaly detection relies on unsupervised time series representation learning. We combine contrastive losses for representation learning with an adversarial learning component that yields environment-invariant embeddings. This environment invariance allows to isolate intrinsic anomalies in the embedding space and leads to favorable performance in the detection tasks. Our empirical evaluation establishes both a framework for evaluating intrinsic anomalies and introduces purposefully designed data sets. We show that the resulting representations allow for meaningful data interactions by exploring nearest neighbors in the embedding space, thereby addressing data quality issues ubiquitous in anomaly detection datasets.

\appendix

\newpage

\section{Additional Dataset Details}\label{app:data_details}

\paragraph{Synthetic data}

We generate two environmental signals ($x_1$ and $x_2$), as well as two system signals ($y_1$, and $y_2$) according to the following set of equations.
\begin{align}
    x_1(x) &= \sin(\frac{x}{275} - 50) + \sin(\frac{x}{200}) + \epsilon_1 \sim \mathcal{N}(0,0.05)\\
    x_2(x) &= \sin(\frac{x}{100})  + \epsilon_2 \sim \mathcal{N}(0,0.1)\\
    y_1(x) &= x_1(x) + \epsilon_3\sim \mathcal{N}(0,0.1)\\
    y_1(x) &= x_1(x) + \frac{x_2(x)}{2} -2 + \epsilon_4\sim \mathcal{N}(0,0.08)
\end{align}

\paragraph{Turbine}

While the turbine dataset consists of 82 distinct time series, our experimentation only operates on a subset of these series. In particular, after preprocessing, we select ambient wind speed (\texttt{Amb\_WindSpeed\_Avg}, \texttt{Amb\_WindSpeed\_Est\_Avg}), ambient wind direction (\texttt{Amb\_WindDir\_Relative\_Avg}, \texttt{Amb\_WindDir\_Abs\_Avg)}, and ambient temperature (\texttt{Amb\_Temp\_Avg}), as our environmental variables and generator rotational speed (\texttt{Gen\_RPM\_Avg}), generator power output (\texttt{Prod\_LatestAvg\_TotActPwr}), and two internal temperature sensors (\texttt{Gear\_Bear\_Temp\_Avg}, \texttt{Hyd\_Oil\_Temp\_Avg}) as our system variables. We select these variables since they do not contain missing values and are easy to interpret. Our core experiments are based on data from the 2016 data set.

\paragraph{Pendulum}
We simulated synthetic data of a pendulum including friction and control signals (rotational acceleration) following the setup in~\citet{kurle2020deep}. 
For simulating the ODE, we used the 4-th order Runge-Kutta method. 
Noisy control signals were sampled from an Ornstein-Uhlenbeck process, where the location of the stationary Gaussian distribution follow a sine signal. 

\section{More Experimental Details}
\label{app:more_exp_details}

\subsection{Hyperparameters}\label{app:hyper_context_inv}

We train all embedding-based methods outlined in Table~\ref{tab:results_auroc} using batches of size 16 and a learning rate of $1.9 \cdot 10^{-3}$ for 50 epochs. The trained TCN encoder model uses 32 channels, a kernel size of 3 and is comprised of 10 causal convolution blocks. The embedding size is dynamically determined based on the length of the time series ($l$) and the number environmental and system dimensions ($d_\text{exog}$ and $d_\text{endog}$) as follows: $l \cdot d_\text{exog} \cdot d_\text{endog} \cdot 0.1$. The negative and positive selection windows mapping into the embedding space are chosen at random locations and are $l \cdot 0.2$ long. Residuals for all residual-based approaches are obtained using the \texttt{MLPRegressor} class from sklearn with out-of-the box parameters. \catchtwotwo\ and \ominanomaly\ are also run using default parameters.

\subsection{Qualitative Results}
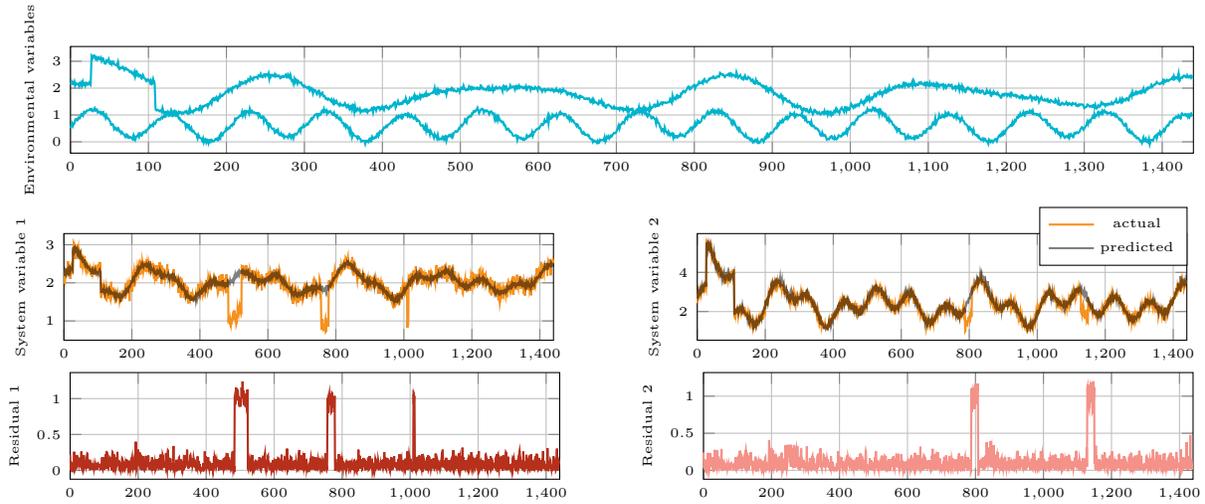
\begin{figure*}
    \centering
    
    \begin{subfigure}[t]{\textwidth}
        \centering
        \begin{tikzpicture}
            \begin{axis}[
              height=3cm,
              width=\linewidth,
              smooth,
              xmin=0,
              xmax=1440,
              grid=major,
              ticklabel style = {font=\tiny},
              label style={font=\tiny},
              ylabel={Environmental variables}
            ]
            
            \addplot [exogencolor, no marks, smooth, thick] table [x=x, y=0, col sep=comma] {figs/syn_exog.csv};
            \addplot [exogencolor, no marks, smooth, thick] table [x=x, y=1, col sep=comma] {figs/syn_exog.csv};
            \end{axis}
        \end{tikzpicture}
    \end{subfigure}
    
    \begin{subfigure}[t]{0.49\textwidth}
        \centering
        \begin{tikzpicture}
            \begin{axis}[
              height=3cm,
              width=\linewidth,
              smooth,
              xmin=0,
              xmax=1440,
              grid=major,
              ticklabel style = {font=\tiny},
              label style={font=\tiny},
              legend style={at={(0.9,0.9)},anchor=north,font=\tiny},
              ylabel={System variable 1}
            ]
            
            \addplot [endogencolor, no marks, smooth, thick] table [x=x, y=0, col sep=comma] {figs/syn_endog.csv};
            \addplot [black, no marks, smooth, opacity=0.5, thick] table [x=x, y=0, col sep=comma] {figs/syn_pred.csv};
            
            \end{axis}
        \end{tikzpicture}
    \end{subfigure}
    ~
    \begin{subfigure}[t]{0.49\textwidth}
        \centering
        \begin{tikzpicture}
            \begin{axis}[
              height=3cm,
              width=\linewidth,
              smooth,
              xmin=0,
              xmax=1440,
              grid=major,
              ticklabel style = {font=\tiny},
              label style={font=\tiny},
              legend style={at={(0.85,1.25)},anchor=north,font=\tiny},
              ylabel={System variable 2}
            ]
            
            \addplot [endogencolor, no marks, smooth, thick] table [x=x, y=1, col sep=comma] {figs/syn_endog.csv};
            \addlegendentry{actual}
            \addplot [black, no marks, smooth, opacity=0.5, thick] table [x=x, y=1, col sep=comma] {figs/syn_pred.csv};
            \addlegendentry{predicted}
            
            \end{axis}
        \end{tikzpicture}
    \end{subfigure}
    
    \begin{subfigure}[t]{0.49\textwidth}
        \centering
        \begin{tikzpicture}
            \begin{axis}[
              height=3cm,
              width=\linewidth,
              smooth,
              xmin=0,
              xmax=1440,
              grid=major,
              ticklabel style = {font=\tiny},
              label style={font=\tiny},
              ylabel={Residual 1}
            ]
            
            \addplot [residualcolor, no marks, smooth, thick] table [x=x, y=0, col sep=comma] {figs/syn_res.csv};
            
            \end{axis}
        \end{tikzpicture}
    \end{subfigure}
    ~
    \begin{subfigure}[t]{0.49\textwidth}
        \centering
        \begin{tikzpicture}
            \begin{axis}[
              height=3cm,
              width=\linewidth,
              smooth,
              xmin=0,
              xmax=1440,
              grid=major,
              ticklabel style = {font=\tiny},
              label style={font=\tiny},
              ylabel={Residual 2}
            ]
            
            \addplot [residualcolor2, no marks, smooth, thick] table [x=x, y=1, col sep=comma] {figs/syn_res.csv};

            \end{axis}
        \end{tikzpicture}
    \end{subfigure}
    \caption{The obtained residuals based on the \synthetic\ dataset. The top panel shows two distinct environmental variables. For the most part, both series follow a smooth seasonal pattern. The first series shows a noticeable extrinsic anomaly at the beginning of the series. In the middle panel, we depict two endogenous variables along with their predictions. Since the environmental variables do give us some information about the presence of an extrinsic anomaly, we can easily predict a similar anomaly for the system variables. However, each system variable exhibits additional anomalies that are deviating from typical expected behaviour. Considering the residuals in the bottom panel, we can see that the residuals only show these intrinsic anomalies but disregard the predicable extrinsic anomaly at the start.}
    \label{fig:syn_residuals}
\end{figure*}

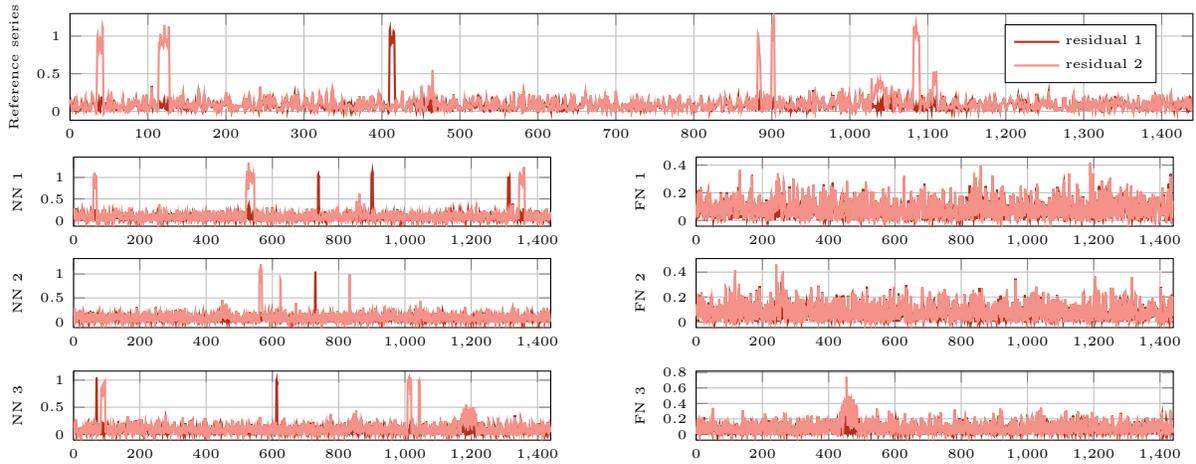
\begin{figure}
    \centering
    
    \begin{subfigure}[t]{\textwidth}
        \centering
        \begin{tikzpicture}
            \begin{axis}[
              height=3cm,
              width=\linewidth,
              smooth,
              xmin=0,
              xmax=1440,
              grid=major,
              ticklabel style = {font=\tiny},
              label style={font=\tiny},
              legend style={at={(0.9,0.9)},anchor=north,font=\tiny},
              ylabel={Reference series}
            ]
            
            \addplot [residualcolor, no marks, smooth, thick] table [x=x, y=0, col sep=comma] {figs/syn_19_ref.csv};
            \addlegendentry{residual 1}
            \addplot [residualcolor2, no marks, smooth, thick] table [x=x, y=1, col sep=comma] {figs/syn_19_ref.csv};
            \addlegendentry{residual 2}
            
            \end{axis}
        \end{tikzpicture}
    \end{subfigure}
    
    \begin{subfigure}[t]{0.48\textwidth}
        \centering
        \begin{tikzpicture}
            \begin{axis}[
              height=2.5cm,
              width=\linewidth,
              smooth,
              xmin=0,
              xmax=1440,
              grid=major,
              ticklabel style = {font=\tiny},
              label style={font=\tiny},
              ylabel={NN 1}
            ]
            
            \addplot [residualcolor, no marks, smooth, thick] table [x=x, y=0, col sep=comma] {figs/syn_19_best_0.csv};
            \addplot [residualcolor2, no marks, smooth, thick] table [x=x, y=0, col sep=comma] {figs/syn_19_best_1.csv};
            \end{axis}
        \end{tikzpicture}
    \end{subfigure}
    ~
    \begin{subfigure}[t]{0.48\textwidth}
        \centering
        \begin{tikzpicture}
            \begin{axis}[
              height=2.5cm,
              width=\linewidth,
              smooth,
              xmin=0,
              xmax=1440,
              grid=major,
              ticklabel style = {font=\tiny},
              label style={font=\tiny},
              ylabel={FN 1}
            ]
            
            \addplot [residualcolor, no marks, smooth, thick] table [x=x, y=0, col sep=comma] {figs/syn_19_worst_0.csv};
            \addplot [residualcolor2, no marks, smooth, thick] table [x=x, y=0, col sep=comma] {figs/syn_19_worst_1.csv};
            \end{axis}
        \end{tikzpicture}
    \end{subfigure}
    
    \begin{subfigure}[t]{0.48\textwidth}
        \centering
        \begin{tikzpicture}
            \begin{axis}[
              height=2.5cm,
              width=\linewidth,
              smooth,
              xmin=0,
              xmax=1440,
              grid=major,
              ticklabel style = {font=\tiny},
              label style={font=\tiny},
              ylabel={NN 2}
            ]
            
            \addplot [residualcolor, no marks, smooth, thick] table [x=x, y=1, col sep=comma] {figs/syn_19_best_0.csv};
            \addplot [residualcolor2, no marks, smooth, thick] table [x=x, y=1, col sep=comma] {figs/syn_19_best_1.csv};
            \end{axis}
        \end{tikzpicture}
    \end{subfigure}
    ~
    \begin{subfigure}[t]{0.48\textwidth}
        \centering
        \begin{tikzpicture}
            \begin{axis}[
              height=2.5cm,
              width=\linewidth,
              smooth,
              xmin=0,
              xmax=1440,
              grid=major,
              ticklabel style = {font=\tiny},
              label style={font=\tiny},
              ylabel={FN 2}
            ]
            
            \addplot [residualcolor, no marks, smooth, thick] table [x=x, y=1, col sep=comma] {figs/syn_19_worst_0.csv};
            \addplot [residualcolor2, no marks, smooth, thick] table [x=x, y=1, col sep=comma] {figs/syn_19_worst_1.csv};
            \end{axis}
        \end{tikzpicture}
    \end{subfigure}
    
    \begin{subfigure}[t]{0.48\textwidth}
        \centering
        \begin{tikzpicture}
            \begin{axis}[
              height=2.5cm,
              width=\linewidth,
              smooth,
              xmin=0,
              xmax=1440,
              grid=major,
              ticklabel style = {font=\tiny},
              label style={font=\tiny},
              ylabel={NN 3}
            ]
            
            \addplot [residualcolor, no marks, smooth, thick] table [x=x, y=2, col sep=comma] {figs/syn_19_best_0.csv};
            \addplot [residualcolor2, no marks, smooth, thick] table [x=x, y=2, col sep=comma] {figs/syn_19_best_1.csv};
            \end{axis}
        \end{tikzpicture}
    \end{subfigure}
    ~
    \begin{subfigure}[t]{0.48\textwidth}
        \centering
        \begin{tikzpicture}
            \begin{axis}[
              height=2.5cm,
              width=\linewidth,
              smooth,
              xmin=0,
              xmax=1440,
              grid=major,
              ticklabel style = {font=\tiny},
              label style={font=\tiny},
              ylabel={FN 3}
            ]
            
            \addplot [residualcolor, no marks, smooth, thick] table [x=x, y=2, col sep=comma] {figs/syn_19_worst_0.csv};
            \addplot [residualcolor2, no marks, smooth, thick] table [x=x, y=2, col sep=comma] {figs/syn_19_worst_1.csv};
            \end{axis}
        \end{tikzpicture}
    \end{subfigure}
    \caption{Nearest-neighbor analysis of \twostepmodelregr\ for the \synthetic\ dataset. The top panel depicts an anomalous reference series showing the two residual series, both of which exhibit anomalous behaviour as indicated by the residual spikes. By considering the nearest (NN) and farthest (FN) neighbour series in the embedding space, we can visually inspect the embedding quality. The left panel shows the 3 nearest neighbour series. Clearly, these series align with the reference series since the same spiking pattern in both residuals is present. Considering the farthest neighbours in the right panel, we see that both residual series are noisy without showing pronounced peaks.}
    \label{fig:syn_res_embedding_1}
\end{figure}

\begin{figure*}
    \centering
    
    \begin{subfigure}[t]{\textwidth}
        \centering
        \begin{tikzpicture}
            \begin{axis}[
              height=3cm,
              width=\linewidth,
              smooth,
              xmin=0,
              xmax=144,
              grid=major,
              ticklabel style = {font=\tiny},
              label style={font=\tiny},
              ylabel={Environmental variables}
            ]
            
            \addplot [exogencolor, no marks, smooth, thick] table [x=x, y=0, col sep=comma] {figs/pend_seq_exog.csv};
            % \addplot [exogencolor, no marks, smooth, thick] table [x=x, y=1, col sep=comma] {figs/pend_seq_exog.csv};
            \end{axis}
        \end{tikzpicture}
    \end{subfigure}
    
    \begin{subfigure}[t]{0.49\textwidth}
        \centering
        \begin{tikzpicture}
            \begin{axis}[
              height=3cm,
              width=\linewidth,
              smooth,
              xmin=0,
              xmax=144,
              grid=major,
              ticklabel style = {font=\tiny},
              label style={font=\tiny},
              legend style={at={(0.2,1.2)},anchor=north,font=\tiny},
              ylabel={System variable 1}
            ]
            
            \addplot [endogencolor, no marks, smooth, thick] table [x=x, y=0, col sep=comma] {figs/pend_seq_endog.csv};
            \addlegendentry{actual}
            \addplot [black, no marks, smooth, opacity=0.5, thick] table [x=x, y=0, col sep=comma] {figs/pend_seq_pred.csv};
            \addlegendentry{predicted}
            
            \end{axis}
        \end{tikzpicture}
    \end{subfigure}
    ~
    \begin{subfigure}[t]{0.49\textwidth}
        \centering
        \begin{tikzpicture}
            \begin{axis}[
              height=3cm,
              width=\linewidth,
              smooth,
              xmin=0,
              xmax=144,
              grid=major,
              ticklabel style = {font=\tiny},
              label style={font=\tiny},
              legend style={at={(0.85,1.25)},anchor=north,font=\tiny},
              ylabel={System variable 2}
            ]
            
            \addplot [endogencolor, no marks, smooth, thick] table [x=x, y=1, col sep=comma] {figs/pend_seq_endog.csv};
            \addplot [black, no marks, smooth, opacity=0.5, thick] table [x=x, y=1, col sep=comma] {figs/pend_seq_pred.csv};
            
            \end{axis}
        \end{tikzpicture}
    \end{subfigure}
    
    \begin{subfigure}[t]{0.49\textwidth}
        \centering
        \begin{tikzpicture}
            \begin{axis}[
              height=3cm,
              width=\linewidth,
              smooth,
              xmin=0,
              xmax=144,
              grid=major,
              ticklabel style = {font=\tiny},
              label style={font=\tiny},
              ylabel={Residual 1}
            ]
            
            \addplot [residualcolor, no marks, smooth, thick] table [x=x, y=0, col sep=comma] {figs/pend_seq_res.csv};
            
            \end{axis}
        \end{tikzpicture}
    \end{subfigure}
    ~
    \begin{subfigure}[t]{0.49\textwidth}
        \centering
        \begin{tikzpicture}
            \begin{axis}[
              height=3cm,
              width=\linewidth,
              smooth,
              xmin=0,
              xmax=144,
              grid=major,
              ticklabel style = {font=\tiny},
              label style={font=\tiny},
              ylabel={Residual 2}
            ]
            
            \addplot [residualcolor2, no marks, smooth, thick] table [x=x, y=1, col sep=comma] {figs/pend_seq_res.csv};

            \end{axis}
        \end{tikzpicture}
    \end{subfigure}
    \caption{The obtained residuals based on the \pendulum\ dataset. For this data set, we show the control signal in blue (top panel) and see that we cannot predict the system variables well with an instant-effect, multi-variate regression.}
    \label{fig:pend_seq_residuals}
\end{figure*}
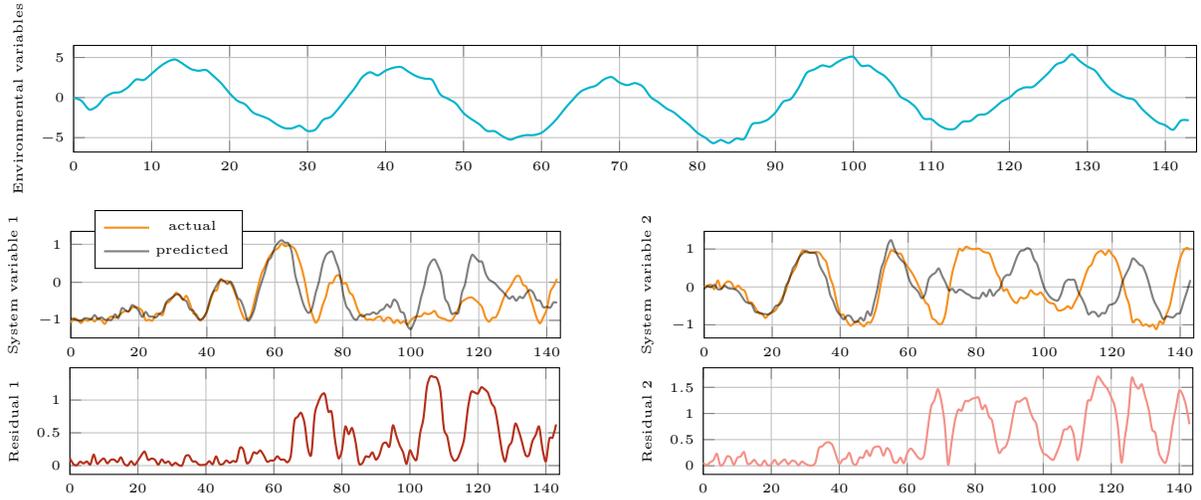

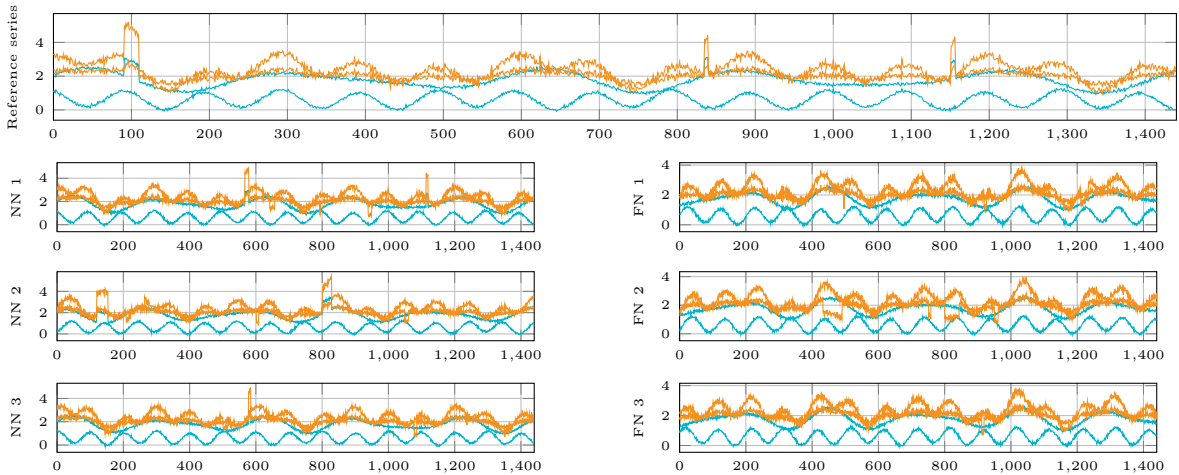
\begin{figure}
    \centering
    
    \begin{subfigure}[t]{\textwidth}
        \centering
        \begin{tikzpicture}
            \begin{axis}[
              height=3cm,
              width=\linewidth,
              smooth,
              xmin=0,
              xmax=1440,
              grid=major,
              ticklabel style = {font=\tiny},
              label style={font=\tiny},
              ylabel={Reference series}
            ]
            
            \addplot [exogencolor, no marks, smooth] table [x=x, y=0, col sep=comma] {figs/full_16_ref.csv};
            \addplot [exogencolor, no marks, smooth] table [x=x, y=1, col sep=comma] {figs/full_16_ref.csv};
            \addplot [endogencolor, no marks, smooth] table [x=x, y=2, col sep=comma] {figs/full_16_ref.csv};
            \addplot [endogencolor, no marks, smooth] table [x=x, y=3, col sep=comma] {figs/full_16_ref.csv};
            
            \end{axis}
        \end{tikzpicture}
    \end{subfigure}
    
    \begin{subfigure}[t]{0.48\textwidth}
        \centering
        \begin{tikzpicture}
            \begin{axis}[
              height=2.5cm,
              width=\linewidth,
              smooth,
              xmin=0,
              xmax=1440,
              grid=major,
              ticklabel style = {font=\tiny},
              label style={font=\tiny},
              ylabel={NN 1}
            ]
            
            \addplot [exogencolor, no marks, smooth] table [x=x, y=0, col sep=comma] {figs/full_16_best_0.csv};
            \addplot [exogencolor, no marks, smooth] table [x=x, y=0, col sep=comma] {figs/full_16_best_1.csv};
            \addplot [endogencolor, no marks, smooth] table [x=x, y=0, col sep=comma] {figs/full_16_best_2.csv};
            \addplot [endogencolor, no marks, smooth] table [x=x, y=0, col sep=comma] {figs/full_16_best_3.csv};
            \end{axis}
        \end{tikzpicture}
    \end{subfigure}
    ~
    \begin{subfigure}[t]{0.48\textwidth}
        \centering
        \begin{tikzpicture}
            \begin{axis}[
              height=2.5cm,
              width=\linewidth,
              smooth,
              xmin=0,
              xmax=1440,
              grid=major,
              ticklabel style = {font=\tiny},
              label style={font=\tiny},
              ylabel={FN 1}
            ]
            
            \addplot [exogencolor, no marks, smooth] table [x=x, y=0, col sep=comma] {figs/full_16_worst_0.csv};
            \addplot [exogencolor, no marks, smooth] table [x=x, y=0, col sep=comma] {figs/full_16_worst_1.csv};
            \addplot [endogencolor, no marks, smooth] table [x=x, y=0, col sep=comma] {figs/full_16_worst_2.csv};
            \addplot [endogencolor, no marks, smooth] table [x=x, y=0, col sep=comma] {figs/full_16_worst_3.csv};
            \end{axis}
        \end{tikzpicture}
    \end{subfigure}
    
    \begin{subfigure}[t]{0.48\textwidth}
        \centering
        \begin{tikzpicture}
            \begin{axis}[
              height=2.5cm,
              width=\linewidth,
              smooth,
              xmin=0,
              xmax=1440,
              grid=major,
              ticklabel style = {font=\tiny},
              label style={font=\tiny},
              ylabel={NN 2}
            ]
            
            \addplot [exogencolor, no marks, smooth] table [x=x, y=1, col sep=comma] {figs/full_16_best_0.csv};
            \addplot [exogencolor, no marks, smooth] table [x=x, y=1, col sep=comma] {figs/full_16_best_1.csv};
            \addplot [endogencolor, no marks, smooth] table [x=x, y=1, col sep=comma] {figs/full_16_best_2.csv};
            \addplot [endogencolor, no marks, smooth] table [x=x, y=1, col sep=comma] {figs/full_16_best_3.csv};
            \end{axis}
        \end{tikzpicture}
    \end{subfigure}
    ~
    \begin{subfigure}[t]{0.48\textwidth}
        \centering
        \begin{tikzpicture}
            \begin{axis}[
              height=2.5cm,
              width=\linewidth,
              smooth,
              xmin=0,
              xmax=1440,
              grid=major,
              ticklabel style = {font=\tiny},
              label style={font=\tiny},
              ylabel={FN 2}
            ]
            
            \addplot [exogencolor, no marks, smooth] table [x=x, y=1, col sep=comma] {figs/full_16_worst_0.csv};
            \addplot [exogencolor, no marks, smooth] table [x=x, y=1, col sep=comma] {figs/full_16_worst_1.csv};
            \addplot [endogencolor, no marks, smooth] table [x=x, y=1, col sep=comma] {figs/full_16_worst_2.csv};
            \addplot [endogencolor, no marks, smooth] table [x=x, y=1, col sep=comma] {figs/full_16_worst_3.csv};
            \end{axis}
        \end{tikzpicture}
    \end{subfigure}
    
    \begin{subfigure}[t]{0.48\textwidth}
        \centering
        \begin{tikzpicture}
            \begin{axis}[
              height=2.5cm,
              width=\linewidth,
              smooth,
              xmin=0,
              xmax=1440,
              grid=major,
              ticklabel style = {font=\tiny},
              label style={font=\tiny},
              ylabel={NN 3}
            ]
            
            \addplot [exogencolor, no marks, smooth] table [x=x, y=2, col sep=comma] {figs/full_16_best_0.csv};
            \addplot [exogencolor, no marks, smooth] table [x=x, y=2, col sep=comma] {figs/full_16_best_1.csv};
            \addplot [endogencolor, no marks, smooth] table [x=x, y=2, col sep=comma] {figs/full_16_best_2.csv};
            \addplot [endogencolor, no marks, smooth] table [x=x, y=2, col sep=comma] {figs/full_16_best_3.csv};
            \end{axis}
        \end{tikzpicture}
    \end{subfigure}
    ~
    \begin{subfigure}[t]{0.48\textwidth}
        \centering
        \begin{tikzpicture}
            \begin{axis}[
              height=2.5cm,
              width=\linewidth,
              smooth,
              xmin=0,
              xmax=1440,
              grid=major,
              ticklabel style = {font=\tiny},
              label style={font=\tiny},
              ylabel={FN 3}
            ]
            
            \addplot [exogencolor, no marks, smooth] table [x=x, y=2, col sep=comma] {figs/full_16_worst_0.csv};
            \addplot [exogencolor, no marks, smooth] table [x=x, y=2, col sep=comma] {figs/full_16_worst_1.csv};
            \addplot [endogencolor, no marks, smooth] table [x=x, y=2, col sep=comma] {figs/full_16_worst_2.csv};
            \addplot [endogencolor, no marks, smooth] table [x=x, y=2, col sep=comma] {figs/full_16_worst_3.csv};
            \end{axis}
        \end{tikzpicture}
    \end{subfigure}
    \caption{Nearest-neighbor analysis of \basicemb\ for the \synthetic\ dataset. The top panel depicts a reference series consisting of both environmental and system signals. The system series do not contain any intrinsic anomalies. In contrast to Figure~\ref{fig:syn_res_embedding_1}, there is no agreement in the nearest neighbors, as the embedding is unable to pick up on the separation into extrinsic anomalies and intrinsic anomalies.}
    \label{fig:syn_full_embedding_1}
\end{figure}

\begin{figure}
    \centering
    
    \begin{subfigure}[t]{\textwidth}
        \centering
        \begin{tikzpicture}
            \begin{axis}[
              height=3cm,
              width=\linewidth,
              smooth,
              xmin=0,
              xmax=144,
              grid=major,
              ticklabel style = {font=\tiny},
              label style={font=\tiny},
              legend style={at={(0.6,0.9)},anchor=north,font=\tiny},
              ylabel={Reference series}
            ]
            
            \addplot [residualcolor, no marks, smooth, thick] table [x=x, y=0, col sep=comma] {figs/16_ref.csv};
            \addlegendentry{rotation residual}
            \addplot [residualcolor2, no marks, smooth, thick] table [x=x, y=1, col sep=comma] {figs/16_ref.csv};
            \addlegendentry{power residual}
            
            \end{axis}
        \end{tikzpicture}
    \end{subfigure}
    
    \begin{subfigure}[t]{0.48\textwidth}
        \centering
        \begin{tikzpicture}
            \begin{axis}[
              height=2.5cm,
              width=\linewidth,
              smooth,
              xmin=0,
              xmax=144,
              grid=major,
              ticklabel style = {font=\tiny},
              label style={font=\tiny},
              ylabel={NN 1}
            ]
            
            \addplot [residualcolor, no marks, smooth, thick] table [x=x, y=0, col sep=comma] {figs/16_best_0.csv};
            \addplot [residualcolor2, no marks, smooth, thick] table [x=x, y=0, col sep=comma] {figs/16_best_1.csv};
            \end{axis}
        \end{tikzpicture}
    \end{subfigure}
    ~
    \begin{subfigure}[t]{0.48\textwidth}
        \centering
        \begin{tikzpicture}
            \begin{axis}[
              height=2.5cm,
              width=\linewidth,
              smooth,
              xmin=0,
              xmax=144,
              grid=major,
              ticklabel style = {font=\tiny},
              label style={font=\tiny},
              ylabel={FN 1}
            ]
            
            \addplot [residualcolor, no marks, smooth, thick] table [x=x, y=0, col sep=comma] {figs/16_worst_0.csv};
            \addplot [residualcolor2, no marks, smooth, thick] table [x=x, y=0, col sep=comma] {figs/16_worst_1.csv};
            \end{axis}
        \end{tikzpicture}
    \end{subfigure}
    
    \begin{subfigure}[t]{0.48\textwidth}
        \centering
        \begin{tikzpicture}
            \begin{axis}[
              height=2.5cm,
              width=\linewidth,
              smooth,
              xmin=0,
              xmax=144,
              grid=major,
              ticklabel style = {font=\tiny},
              label style={font=\tiny},
              ylabel={NN 2}
            ]
            
            \addplot [residualcolor, no marks, smooth, thick] table [x=x, y=1, col sep=comma] {figs/16_best_0.csv};
            \addplot [residualcolor2, no marks, smooth, thick] table [x=x, y=1, col sep=comma] {figs/16_best_1.csv};
            \end{axis}
        \end{tikzpicture}
    \end{subfigure}
    ~
    \begin{subfigure}[t]{0.48\textwidth}
        \centering
        \begin{tikzpicture}
            \begin{axis}[
              height=2.5cm,
              width=\linewidth,
              smooth,
              xmin=0,
              xmax=144,
              grid=major,
              ticklabel style = {font=\tiny},
              label style={font=\tiny},
              ylabel={FN 2}
            ]
            
            \addplot [residualcolor, no marks, smooth, thick] table [x=x, y=1, col sep=comma] {figs/16_worst_0.csv};
            \addplot [residualcolor2, no marks, smooth, thick] table [x=x, y=1, col sep=comma] {figs/16_worst_1.csv};
            \end{axis}
        \end{tikzpicture}
    \end{subfigure}
    
    \begin{subfigure}[t]{0.48\textwidth}
        \centering
        \begin{tikzpicture}
            \begin{axis}[
              height=2.5cm,
              width=\linewidth,
              smooth,
              xmin=0,
              xmax=144,
              grid=major,
              ticklabel style = {font=\tiny},
              label style={font=\tiny},
              ylabel={NN 3}
            ]
            
            \addplot [residualcolor, no marks, smooth, thick] table [x=x, y=2, col sep=comma] {figs/16_best_0.csv};
            \addplot [residualcolor2, no marks, smooth, thick] table [x=x, y=2, col sep=comma] {figs/16_best_1.csv};
            \end{axis}
        \end{tikzpicture}
    \end{subfigure}
    ~
    \begin{subfigure}[t]{0.48\textwidth}
        \centering
        \begin{tikzpicture}
            \begin{axis}[
              height=2.5cm,
              width=\linewidth,
              smooth,
              xmin=0,
              xmax=144,
              grid=major,
              ticklabel style = {font=\tiny},
              label style={font=\tiny},
              ylabel={FN 3}
            ]
            
            \addplot [residualcolor, no marks, smooth, thick] table [x=x, y=2, col sep=comma] {figs/16_worst_0.csv};
            \addplot [residualcolor2, no marks, smooth, thick] table [x=x, y=2, col sep=comma] {figs/16_worst_1.csv};
            \end{axis}
        \end{tikzpicture}
    \end{subfigure}
    \caption{Nearest-neighbor analysis of \twostepmodelregr\ for the \turbine\ dataset. Similar to Figure~\ref{fig:syn_res_embedding_1}}
    \label{fig:turb_res_embedding_1}
\end{figure}
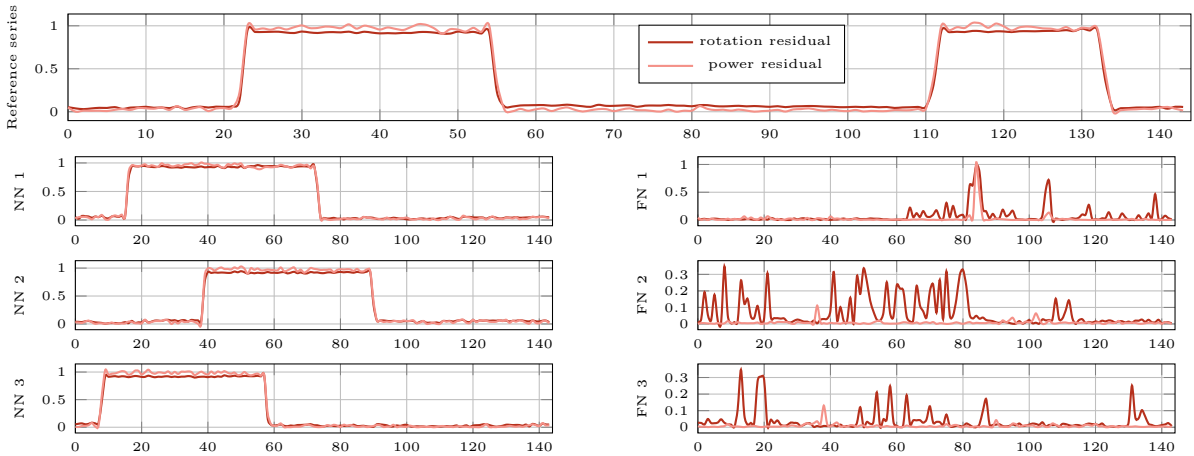

Figure~\ref{fig:syn_residuals} depicts the effect of isolating intrinsic anomalies via a predictive network on \synthetic\ data. In Figures~\ref{fig:syn_res_embedding_1}, we show the effect of learning embeddings on the residuals vs on the actual values. Figure~\ref{fig:syn_full_embedding_1} shows qualitatively that we need to modify off-the-shelf time series embeddings in order to be able to isolate intrinsic anomalies. Similar results are available also for the \turbine dataset in Figures~\ref{fig:turb_residuals} and \ref{fig:turb_res_embedding_1}. Figure~\ref{fig:pend_seq_residuals} show that purely residual-based approaches can fail in more complex prediction scenarios.

%%!TEX root = ../rewrite.tex

%%!TEX root = ../rewrite.tex

%%!TEX root = ../rewrite.tex

%%!TEX root = ../rewrite.tex

\begin{figure}
    \centering
    
    \begin{subfigure}[t]{\textwidth}
        \centering
        \begin{tikzpicture}
            \begin{axis}[
              height=3.2cm,
              width=\linewidth,
              smooth,
              xmin=0,
              xmax=1400,
              grid=major,
              ticklabel style = {font=\tiny},
              label style={font=\tiny},
              ylabel={Environmental variables}
            ]
            
            \addplot [exogencolor, no marks, smooth, thick] table [x=x, y=0, col sep=comma] {figs/turb_exog.csv};
            \addplot [exogencolor, no marks, smooth, thick] table [x=x, y=1, col sep=comma] {figs/turb_exog.csv};
            \addplot [exogencolor, no marks, smooth, thick] table [x=x, y=2, col sep=comma] {figs/turb_exog.csv};
            \addplot [exogencolor, no marks, smooth, thick] table [x=x, y=3, col sep=comma] {figs/turb_exog.csv};
            \addplot [exogencolor, no marks, smooth, thick] table [x=x, y=4, col sep=comma] {figs/turb_exog.csv};
            
            \end{axis}
        \end{tikzpicture}
    \end{subfigure}
    
    \begin{subfigure}[t]{0.49\textwidth}
        \centering
        \begin{tikzpicture}
            \begin{axis}[
              height=3.2cm,
              width=\textwidth,
              smooth,
              xmin=0,
              xmax=1400,
              grid=major,
              ticklabel style = {font=\tiny},
              label style={font=\tiny},
              legend style={at={(0.68,1.1)},anchor=north,font=\tiny},
              ylabel={Power}
            ]
            
            \addplot [endogencolor, no marks, smooth, thick] table [x=x, y=1, col sep=comma] {figs/turb_endog.csv};
            \addlegendentry{actual}
            \addplot [black, no marks, smooth, opacity=0.5, thick] table [x=x, y=1, col sep=comma] {figs/turb_pred.csv};
            \addlegendentry{predicted}
%            \addplot [endogencolor, no marks, smooth] table [x=x, y=1, col sep=comma] {figs/syn_endog.csv};
            
            \end{axis}
        \end{tikzpicture}
    \end{subfigure}
    ~
    \begin{subfigure}[t]{0.49\textwidth}
        \centering
        \begin{tikzpicture}
            \begin{axis}[
              height=3.2cm,
              width=\textwidth,
              smooth,
              xmin=0,
              xmax=1400,
              grid=major,
              ticklabel style = {font=\tiny},
              label style={font=\tiny},
              ylabel={Rotation}
            ]
            
            \addplot [endogencolor, no marks, smooth, thick] table [x=x, y=0, col sep=comma] {figs/turb_endog.csv};
            \addplot [black, no marks, smooth, opacity=0.5, thick] table [x=x, y=0, col sep=comma] {figs/turb_pred.csv};
            
            \end{axis}
        \end{tikzpicture}
    \end{subfigure}
    
    \begin{subfigure}[t]{0.49\textwidth}
        \centering
        \begin{tikzpicture}
            \begin{axis}[
              height=3.2cm,
              width=\linewidth,
              smooth,
              xmin=0,
              xmax=1400,
              grid=major,
              ticklabel style = {font=\tiny},
              label style={font=\tiny},
              ylabel={Power Residual}
            ]
            
            \addplot [residualcolor, no marks, smooth, thick] table [x=x, y=1, col sep=comma] {figs/turb_res.csv};
            
            \end{axis}
        \end{tikzpicture}
    \end{subfigure}
    ~
    \begin{subfigure}[t]{0.49\textwidth}
        \centering
        \begin{tikzpicture}
            \begin{axis}[
              height=3.2cm,
              width=\linewidth,
              smooth,
              xmin=0,
              xmax=1400,
              grid=major,
              ticklabel style = {font=\tiny},
              label style={font=\tiny},
              ylabel={Rotation Residual}
            ]
            
            \addplot [residualcolor, no marks, smooth, thick] table [x=x, y=0, col sep=comma] {figs/turb_res.csv};
            
            \end{axis}
        \end{tikzpicture}
    \end{subfigure}
    \caption{The obtained residuals based on the \turbine\ dataset. Similar to Figure~\ref{fig:syn_residuals}.}
    \label{fig:turb_residuals}
\end{figure}
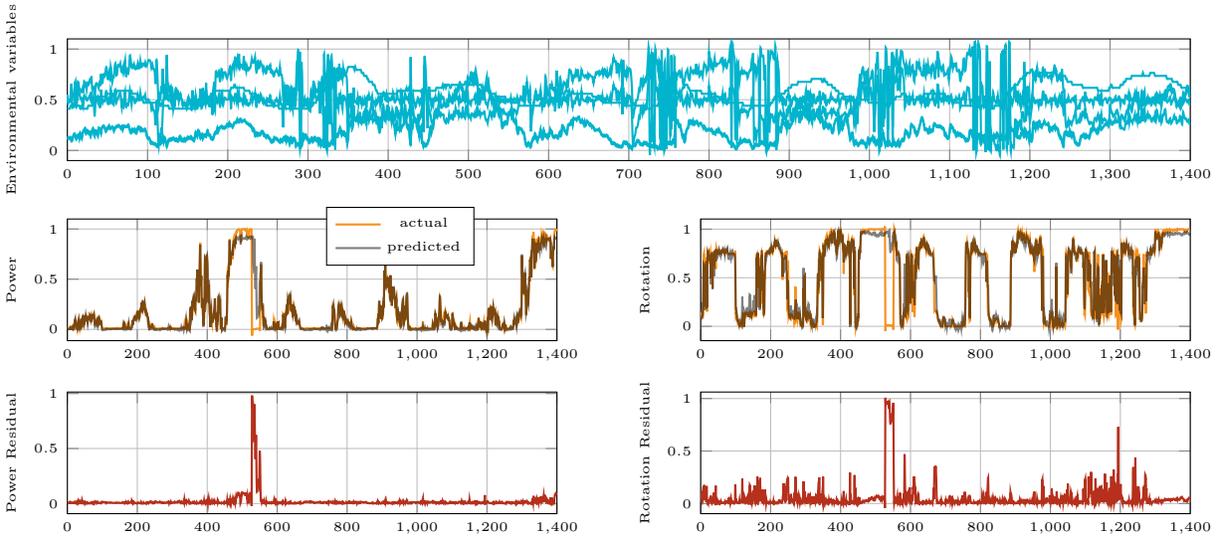

\subsection{Quantitative Results}

\paragraph{Embedding Space Distances} We provide more fine-grained results for the embedding space distance experiment in Table~\ref{tab:results_distance_ext}.

\paragraph{The importance of $\lambda$} We perform an ablation study to determine the influence of trading off the contrastive loss with the adversarial environment-invariance loss in~\shorteqref{eq:combined-loss}, denoted by $\lambda$. The results in Table~\ref{tab:lambda_abl}, demonstrate that (i) our particular choice of negative examples already provide a strong degree of environment invariance ($\lambda = 0$); and that (ii) these results can be further improved by a careful weighting of the adversarial component. Overall, $\lambda = 10^{-3}$ gives the strongest results, which we also use in our main experimental results in Table~\ref{tab:results_auroc}.

\begin{table*}[]
    \centering
    \small
    \setlength{\tabcolsep}{3.5pt}
    \caption{Sensitivity study on the \synthetic \ dataset for \domainadv. Dependency-breaking negative examples alone $(\lambda=0)$ already provide strong performance and adding adversarial training with a carefully chosen $\lambda = 10^{-3}$ further improves results.}
    \begin{tabular}{cccccccc}
        \toprule
        $\lambda$ & 0 & $10^{-5}$ & $10^{-4}$ & $10^{-3}$ & $10^{-2}$ & $10^{-1}$ & 1\\
    \midrule
        AUROC & \textbf{0.85} ($\pm$ 0.12) & 0.69 ($\pm$ 0.15) & \textbf{0.86} ($\pm$ 0.06) & \textbf{0.94} ($\pm$ 0.03) & \textbf{0.84} ($\pm$ 0.14) & 0.58 ($\pm$ 0.09) & 0.55 ($\pm$ 0.07)\\
        F1 & \textbf{0.96} ($\pm$ 0.03)  & 0.92 ($\pm$ 0.04) & 0.96 ($\pm$ 0.02) & \textbf{0.99} ($\pm$ 0.01) & 0.97 ($\pm$ 0.03) & 0.87 ($\pm$ 0.04) & 0.86 ($\pm$ 0.04) \\
    \bottomrule
    \end{tabular}
    \label{tab:lambda_abl}
\end{table*}

\begin{sidewaystable}
  \centering
  \vspace{5pt}
  \scriptsize
  \caption{Distance results on the anomaly detection task with 5 random seeds. Distances to the correct class (Corr) should be small while distances to the incorrect class (Incorr) should be large. We prefer approaches with a large gap between the mean correct disatance and the mean incorrect distance.}
  \begin{tabular}{cccccccccccc}
    \toprule
    & \multicolumn{3}{c}{\basicemb} & \multicolumn{3}{c}{\twostepmodelregr} & \multicolumn{3}{c}{\domainadv} \\
    \cmidrule(r){2-4}
    \cmidrule(r){5-7}
    \cmidrule(r){8-10}
    Dataset & Corr & Incorr & Gap & Corr & Incorr & Gap & Corr & Incorr & Gap\\
    \midrule
    \synthetic & 0.483 ($\pm$ 0.249) & 0.534 ($\pm$ 0.257) & 0.051 ($\pm$ 0.041) & 0.634 ($\pm$ 0.216) & 1.284 ($\pm$ 0.122) & 0.641 ($\pm$ 0.314) & 0.034 ($\pm$ 0.012) & 1.978 ($\pm$ 0.029) & 1.944 ($\pm$ 0.039)\\
    \devops & 0.918 ($\pm$ 0.038) & 0.932 ($\pm$ 0.034) & 0.446 ($\pm$ 0.059) & 0.942 ($\pm$ 0.016) & 0.977 ($\pm$ 0.020) & 0.433 ($\pm$ 0.074) & 0.043 ($\pm$ 0.133) & 0.529 ($\pm$ 0.130) & 0.486 ($\pm$ 0.178)\\
    \pendulum & 0.691 ($\pm$ 0.037) & 1.137 ($\pm$ 0.076) & 0.014 ($\pm$ 0.006) & 0.679 ($\pm$ 0.036) & 1.112 ($\pm$ 0.065) & 0.035 ($\pm$ 0.008) & 0.215 ($\pm$ 0.178) & 0.437 ($\pm$ 0.346) & 0.012 ($\pm$ 0.007)\\
    \turbine & 0.927 ($\pm$ 0.024) & 1.002 ($\pm$ 0.022) & 0.075 ($\pm$ 0.007) & 0.691 ($\pm$ 0.186) & 1.010 ($\pm$ 0.272) & 0.319 ($\pm$ 0.103) & 0.065 ($\pm$ 0.079) & 0.516 ($\pm$ 0.161) & 0.450 ($\pm$ 0.085)\\
    \bottomrule
  \end{tabular}
  \label{tab:results_distance_ext}
\end{sidewaystable}


\newpage

\begin{thebibliography}{47}
\providecommand{\natexlab}[1]{#1}
\providecommand{\url}[1]{\texttt{#1}}
\expandafter\ifx\csname urlstyle\endcsname\relax
  \providecommand{\doi}[1]{doi: #1}\else
  \providecommand{\doi}{doi: \begingroup \urlstyle{rm}\Url}\fi

\bibitem[Achille and Soatto(2018)]{achille18}
Alessandro Achille and Stefano Soatto.
\newblock Emergence of invariance and disentanglement in deep representations.
\newblock \emph{Journal of Machine Learning Research}, 19\penalty0
  (50):\penalty0 1--34, 2018.
\newblock URL \url{http://jmlr.org/papers/v19/17-646.html}.

\bibitem[Akash et~al.(2021)Akash, Lokhande, Ravi, and Singh]{akash2021learning}
Aditya~Kumar Akash, Vishnu~Suresh Lokhande, Sathya~N. Ravi, and Vikas Singh.
\newblock Learning invariant representations using inverse contrastive loss,
  2021.

\bibitem[Ayed et~al.(2020)Ayed, Stella, Januschowski, and
  Gasthaus]{ayed2020anomaly}
Fadhel Ayed, Lorenzo Stella, Tim Januschowski, and Jan Gasthaus.
\newblock Anomaly detection at scale: The case for deep distributional time
  series models, 2020.

\bibitem[Bai et~al.(2018)Bai, Kolter, and Koltun]{tcn}
Shaojie Bai, J.~Zico Kolter, and Vladlen Koltun.
\newblock An empirical evaluation of generic convolutional and recurrent
  networks for sequence modeling, 2018.

\bibitem[Bl\'{a}zquez-Garc\'{\i}a et~al.(2021)Bl\'{a}zquez-Garc\'{\i}a, Conde,
  Mori, and Lozano]{blazquezgarcia2020review}
Ane Bl\'{a}zquez-Garc\'{\i}a, Angel Conde, Usue Mori, and Jose~A. Lozano.
\newblock A review on outlier/anomaly detection in time series data.
\newblock \emph{ACM Comput. Surv.}, 54\penalty0 (3), apr 2021.
\newblock ISSN 0360-0300.
\newblock \doi{10.1145/3444690}.
\newblock URL \url{https://doi.org/10.1145/3444690}.

\bibitem[Bogatinovski et~al.(2021)Bogatinovski, Nedelkoski, Acker, Schmidt,
  Wittkopp, Becker, Cardoso, and Kao]{bogatinovski2021artificial}
Jasmin Bogatinovski, Sasho Nedelkoski, Alexander Acker, Florian Schmidt,
  Thorsten Wittkopp, Soeren Becker, Jorge Cardoso, and Odej Kao.
\newblock Artificial intelligence for it operations (aiops) workshop white
  paper, 2021.

\bibitem[Breunig et~al.(2000)Breunig, Kriegel, Ng, and Sander]{breunig2000lof}
Markus~M Breunig, Hans-Peter Kriegel, Raymond~T Ng, and J{\"o}rg Sander.
\newblock Lof: identifying density-based local outliers.
\newblock In \emph{Proceedings of the 2000 ACM SIGMOD international conference
  on Management of data}, pages 93--104, 2000.

\bibitem[Carmona et~al.(2021)Carmona, Aubet, Flunkert, and
  Gasthaus]{carmona2021neural}
Chris~U. Carmona, François-Xavier Aubet, Valentin Flunkert, and Jan Gasthaus.
\newblock Neural contextual anomaly detection for time series, 2021.

\bibitem[Chen et~al.(2020)Chen, Kornblith, Norouzi, and Hinton]{chen2020simple}
Ting Chen, Simon Kornblith, Mohammad Norouzi, and Geoffrey Hinton.
\newblock A simple framework for contrastive learning of visual
  representations.
\newblock In \emph{International conference on machine learning}, pages
  1597--1607. PMLR, 2020.

\bibitem[Christ et~al.(2017)Christ, Kempa-Liehr, and Feindt]{tsfresh}
Maximilian Christ, Andreas~W. Kempa-Liehr, and Michael Feindt.
\newblock Distributed and parallel time series feature extraction for
  industrial big data applications, 2017.

\bibitem[Fang et~al.(2020)Fang, Wang, Zhou, Ding, and Xie]{fang2020cert}
Hongchao Fang, Sicheng Wang, Meng Zhou, Jiayuan Ding, and Pengtao Xie.
\newblock Cert: Contrastive self-supervised learning for language
  understanding, 2020.

\bibitem[Foorthuis(2021)]{nature_anomalies_21}
Ralph Foorthuis.
\newblock On the nature and types of anomalies: a review of deviations in data.
\newblock \emph{International Journal of Data Science and Analytics},
  12\penalty0 (4):\penalty0 297–331, Aug 2021.
\newblock ISSN 2364-4168.
\newblock \doi{10.1007/s41060-021-00265-1}.
\newblock URL \url{http://dx.doi.org/10.1007/s41060-021-00265-1}.

\bibitem[Franceschi et~al.(2019)Franceschi, Dieuleveut, and
  Jaggi]{franceschi2019unsupervised}
Jean-Yves Franceschi, Aymeric Dieuleveut, and Martin Jaggi.
\newblock Unsupervised scalable representation learning for multivariate time
  series.
\newblock \emph{arXiv preprint arXiv:1901.10738}, 2019.

\bibitem[Ganin et~al.(2016)Ganin, Ustinova, Ajakan, Germain, Larochelle,
  Laviolette, Marchand, and Lempitsky]{ganin2016domain}
Yaroslav Ganin, Evgeniya Ustinova, Hana Ajakan, Pascal Germain, Hugo
  Larochelle, Fran{\c{c}}ois Laviolette, Mario Marchand, and Victor Lempitsky.
\newblock Domain-adversarial training of neural networks.
\newblock \emph{The journal of machine learning research}, 17\penalty0
  (1):\penalty0 2096--2030, 2016.

\bibitem[Guha et~al.(2016)Guha, Mishra, Roy, and Schrijvers]{guha16}
Sudipto Guha, Nina Mishra, Gourav Roy, and Okke Schrijvers.
\newblock Robust random cut forest based anomaly detection on streams.
\newblock In \emph{Proceedings of the 33rd International Conference on
  International Conference on Machine Learning - Volume 48}, ICML'16, page
  2712–2721. JMLR.org, 2016.

\bibitem[Han et~al.(2012)Han, Kamber, and Pei]{han2012mining}
Jiawei Han, Micheline Kamber, and Jian Pei.
\newblock Data mining concepts and techniques, third edition, 2012.
\newblock URL
  \url{http://www.amazon.de/Data-Mining-Concepts-Techniques-Management/dp/0123814790/ref=tmm_hrd_title_0?ie=UTF8&qid=1366039033&sr=1-1}.

\bibitem[Haufe et~al.(2009)Haufe, Nolte, Mueller, and Kraemer]{haufe2009sparse}
Stefan Haufe, Guido Nolte, Klaus-Robert Mueller, and Nicole Kraemer.
\newblock Sparse causal discovery in multivariate time series, 2009.

\bibitem[He et~al.(2020)He, Fan, Wu, Xie, and Girshick]{he2020momentum}
Kaiming He, Haoqi Fan, Yuxin Wu, Saining Xie, and Ross Girshick.
\newblock Momentum contrast for unsupervised visual representation learning,
  2020.

\bibitem[Jaiswal et~al.(2021)Jaiswal, Babu, Zadeh, Banerjee, and
  Makedon]{jaiswal2021survey}
Ashish Jaiswal, Ashwin~Ramesh Babu, Mohammad~Zaki Zadeh, Debapriya Banerjee,
  and Fillia Makedon.
\newblock A survey on contrastive self-supervised learning, 2021.

\bibitem[Janzing et~al.(2019)Janzing, Budhathoki, Minorics, and
  Blöbaum]{janzing2019causal}
Dominik Janzing, Kailash Budhathoki, Lenon Minorics, and Patrick Blöbaum.
\newblock Causal structure based root cause analysis of outliers, 2019.

\bibitem[Jin et~al.(2021)Jin, Park, Maddix, Wang, and Yan]{jin2021domain}
Xiaoyong Jin, Youngsuk Park, Danielle~C. Maddix, Yuyang Wang, and Xifeng Yan.
\newblock Domain adaptation for time series forecasting via attention sharing,
  2021.

\bibitem[Kim et~al.(2021)Kim, Choi, Choi, Lee, and Yoon]{kim2021towards}
Siwon Kim, Kukjin Choi, Hyun-Soo Choi, Byunghan Lee, and Sungroh Yoon.
\newblock Towards a rigorous evaluation of time-series anomaly detection.
\newblock \emph{arXiv preprint arXiv:2109.05257}, 2021.

\bibitem[Krupitzer et~al.(2020)Krupitzer, Wagenhals, Z{\"u}fle, Lesch,
  Sch{\"a}fer, Mozaffarin, Edinger, Becker, and Kounev]{krupitzer2020survey}
Christian Krupitzer, Tim Wagenhals, Marwin Z{\"u}fle, Veronika Lesch, Dominik
  Sch{\"a}fer, Amin Mozaffarin, Janick Edinger, Christian Becker, and Samuel
  Kounev.
\newblock A survey on predictive maintenance for industry 4.0.
\newblock \emph{arXiv preprint arXiv:2002.08224}, 2020.

\bibitem[Kurle et~al.(2020)Kurle, Rangapuram, de~B{\'e}zenac, G{\"u}nnemann,
  and Gasthaus]{kurle2020deep}
Richard Kurle, Syama~Sundar Rangapuram, Emmanuel de~B{\'e}zenac, Stephan
  G{\"u}nnemann, and Jan Gasthaus.
\newblock Deep rao-blackwellised particle filters for time series forecasting.
\newblock \emph{Advances in Neural Information Processing Systems}, 33, 2020.

\bibitem[Li et~al.(2020)Li, Jiang, Li, Hassan, He, Huang, Zeng, Wang, and
  Chen]{aiops}
Yangguang Li, Zhen Ming~(Jack) Jiang, Heng Li, Ahmed~E. Hassan, Cheng He,
  Ruirui Huang, Zhengda Zeng, Mian Wang, and Pinan Chen.
\newblock Predicting node failures in an ultra-large-scale cloud computing
  platform: An aiops solution.
\newblock \emph{ACM Trans. Softw. Eng. Methodol.}, 29\penalty0 (2), April 2020.
\newblock ISSN 1049-331X.
\newblock \doi{10.1145/3385187}.
\newblock URL \url{https://doi.org/10.1145/3385187}.

\bibitem[Liu et~al.(2008)Liu, Ting, and Zhou]{liu2008isolation}
Fei~Tony Liu, Kai~Ming Ting, and Zhi-Hua Zhou.
\newblock Isolation forest.
\newblock In \emph{2008 eighth ieee international conference on data mining},
  pages 413--422. IEEE, 2008.

\bibitem[Lohrmann and Kao(2011)]{smartmeter}
Björn Lohrmann and Odej Kao.
\newblock Processing smart meter data streams in the cloud.
\newblock In \emph{2011 2nd IEEE PES International Conference and Exhibition on
  Innovative Smart Grid Technologies}, pages 1--8, 2011.
\newblock \doi{10.1109/ISGTEurope.2011.6162747}.

\bibitem[Long et~al.(2017)Long, Cao, Wang, and Jordan]{long2017conditional}
Mingsheng Long, Zhangjie Cao, Jianmin Wang, and Michael~I Jordan.
\newblock Conditional adversarial domain adaptation.
\newblock \emph{arXiv preprint arXiv:1705.10667}, 2017.

\bibitem[Lu et~al.(2009)Lu, Li, Wu, and Yang]{turbines}
Bin Lu, Yaoyu Li, Xin Wu, and Zhongzhou Yang.
\newblock A review of recent advances in wind turbine condition monitoring and
  fault diagnosis.
\newblock In \emph{2009 IEEE Power Electronics and Machines in Wind
  Applications}, pages 1--7, 2009.
\newblock \doi{10.1109/PEMWA.2009.5208325}.

\bibitem[Lubba et~al.(2019)Lubba, Sethi, Knaute, Schultz, Fulcher, and
  Jones]{lubba2019catch22}
Carl~H Lubba, Sarab~S Sethi, Philip Knaute, Simon~R Schultz, Ben~D Fulcher, and
  Nick~S Jones.
\newblock catch22: Canonical time-series characteristics.
\newblock \emph{Data Mining and Knowledge Discovery}, 33\penalty0 (6):\penalty0
  1821--1852, 2019.

\bibitem[Moyer et~al.(2018)Moyer, Gao, Brekelmans, Steeg, and
  Galstyan]{moyer18}
Daniel Moyer, Shuyang Gao, Rob Brekelmans, Greg~Ver Steeg, and Aram Galstyan.
\newblock Invariant representations without adversarial training.
\newblock In \emph{Proceedings of the 32nd International Conference on Neural
  Information Processing Systems}, NIPS'18, page 9102–9111, Red Hook, NY,
  USA, 2018. Curran Associates Inc.

\bibitem[Nedelkoski et~al.(2019)Nedelkoski, Cardoso, and Kao]{8814585}
Sasho Nedelkoski, Jorge Cardoso, and Odej Kao.
\newblock Anomaly detection from system tracing data using multimodal deep
  learning.
\newblock In \emph{2019 IEEE 12th International Conference on Cloud Computing
  (CLOUD)}, pages 179--186, 2019.
\newblock \doi{10.1109/CLOUD.2019.00038}.

\bibitem[Qiu et~al.(2012)Qiu, Liu, Subrahmanya, and Li]{qiu2012granger}
Huida Qiu, Yan Liu, Niranjan~A Subrahmanya, and Weichang Li.
\newblock Granger causality for time-series anomaly detection.
\newblock In \emph{2012 IEEE 12th international conference on data mining},
  pages 1074--1079. IEEE, 2012.

\bibitem[Qiu et~al.(2020)Qiu, Du, Yin, Zhang, and Qian]{app10062166}
Juan Qiu, Qingfeng Du, Kanglin Yin, Shuang-Li Zhang, and Chongshu Qian.
\newblock A causality mining and knowledge graph based method of root cause
  diagnosis for performance anomaly in cloud applications.
\newblock \emph{Applied Sciences}, 10\penalty0 (6), 2020.
\newblock ISSN 2076-3417.
\newblock \doi{10.3390/app10062166}.
\newblock URL \url{https://www.mdpi.com/2076-3417/10/6/2166}.

\bibitem[Ren et~al.(2019)Ren, Xu, Wang, Yi, Huang, Kou, Xing, Yang, Tong, and
  Zhang]{ren19}
Hansheng Ren, Bixiong Xu, Yujing Wang, Chao Yi, Congrui Huang, Xiaoyu Kou, Tony
  Xing, Mao Yang, Jie Tong, and Qi~Zhang.
\newblock Time-series anomaly detection service at microsoft.
\newblock In \emph{Proceedings of the 25th ACM SIGKDD International Conference
  on Knowledge Discovery \& Data Mining}, KDD '19, page 3009–3017, New York,
  NY, USA, 2019. Association for Computing Machinery.
\newblock ISBN 9781450362016.
\newblock \doi{10.1145/3292500.3330680}.
\newblock URL \url{https://doi.org/10.1145/3292500.3330680}.

\bibitem[Ruff et~al.(2018)Ruff, Vandermeulen, Goernitz, Deecke, Siddiqui,
  Binder, M{\"u}ller, and Kloft]{ruff18a}
Lukas Ruff, Robert Vandermeulen, Nico Goernitz, Lucas Deecke, Shoaib~Ahmed
  Siddiqui, Alexander Binder, Emmanuel M{\"u}ller, and Marius Kloft.
\newblock Deep one-class classification.
\newblock In Jennifer Dy and Andreas Krause, editors, \emph{Proceedings of the
  35th International Conference on Machine Learning}, volume~80 of
  \emph{Proceedings of Machine Learning Research}, pages 4393--4402. PMLR,
  10--15 Jul 2018.
\newblock URL \url{https://proceedings.mlr.press/v80/ruff18a.html}.

\bibitem[Sch{\"o}lkopf et~al.(1999)Sch{\"o}lkopf, Williamson, Smola,
  Shawe-Taylor, Platt, et~al.]{scholkopf1999support}
Bernhard Sch{\"o}lkopf, Robert~C Williamson, Alexander~J Smola, John
  Shawe-Taylor, John~C Platt, et~al.
\newblock Support vector method for novelty detection.
\newblock In \emph{NIPS}, volume~12, pages 582--588. Citeseer, 1999.

\bibitem[Sohn et~al.(2021)Sohn, Li, Yoon, Jin, and Pfister]{sohn2021learning}
Kihyuk Sohn, Chun-Liang Li, Jinsung Yoon, Minho Jin, and Tomas Pfister.
\newblock Learning and evaluating representations for deep one-class
  classification, 2021.

\bibitem[Song et~al.(2007)Song, Wu, Jermaine, and Ranka]{4138201}
Xiuyao Song, Mingxi Wu, Christopher Jermaine, and Sanjay Ranka.
\newblock Conditional anomaly detection.
\newblock \emph{IEEE Transactions on Knowledge and Data Engineering},
  19\penalty0 (5):\penalty0 631--645, 2007.
\newblock \doi{10.1109/TKDE.2007.1009}.

\bibitem[Su et~al.(2019{\natexlab{a}})Su, Zhao, Niu, Liu, Sun, and
  Pei]{omnianomaly}
Ya~Su, Youjian Zhao, Chenhao Niu, Rong Liu, Wei Sun, and Dan Pei.
\newblock Robust anomaly detection for multivariate time series through
  stochastic recurrent neural network.
\newblock In \emph{Proceedings of the 25th ACM SIGKDD International Conference
  on Knowledge Discovery \& Data Mining}, KDD '19, page 2828–2837, New York,
  NY, USA, 2019{\natexlab{a}}. Association for Computing Machinery.
\newblock ISBN 9781450362016.
\newblock \doi{10.1145/3292500.3330672}.
\newblock URL \url{https://doi.org/10.1145/3292500.3330672}.

\bibitem[Su et~al.(2019{\natexlab{b}})Su, Zhao, Niu, Liu, Sun, and Pei]{ya19}
Ya~Su, Youjian Zhao, Chenhao Niu, Rong Liu, Wei Sun, and Dan Pei.
\newblock Robust anomaly detection for multivariate time series through
  stochastic recurrent neural network.
\newblock In \emph{Proceedings of the 25th ACM SIGKDD International Conference
  on Knowledge Discovery \& Data Mining}, KDD '19, page 2828–2837, New York,
  NY, USA, 2019{\natexlab{b}}. Association for Computing Machinery.
\newblock ISBN 9781450362016.
\newblock \doi{10.1145/3292500.3330672}.
\newblock URL \url{https://doi.org/10.1145/3292500.3330672}.

\bibitem[Tack et~al.(2020)Tack, Mo, Jeong, and Shin]{NEURIPS2020_8965f766}
Jihoon Tack, Sangwoo Mo, Jongheon Jeong, and Jinwoo Shin.
\newblock Csi: Novelty detection via contrastive learning on distributionally
  shifted instances.
\newblock In H.~Larochelle, M.~Ranzato, R.~Hadsell, M.~F. Balcan, and H.~Lin,
  editors, \emph{Advances in Neural Information Processing Systems}, volume~33,
  pages 11839--11852. Curran Associates, Inc., 2020.
\newblock URL
  \url{https://proceedings.neurips.cc/paper/2020/file/8965f76632d7672e7d3cf29c87ecaa0c-Paper.pdf}.

\bibitem[van~den Oord et~al.(2019)van~den Oord, Li, and
  Vinyals]{oord2019representation}
Aaron van~den Oord, Yazhe Li, and Oriol Vinyals.
\newblock Representation learning with contrastive predictive coding, 2019.

\bibitem[Wu et~al.(2020)Wu, Bogatinovski, Nedelkoski, Tordsson, and
  Kao]{sockshop}
Li~Wu, Jasmin Bogatinovski, Sasho Nedelkoski, Johan Tordsson, and Odej Kao.
\newblock {Performance Diagnosis in Cloud Microservices using Deep Learning}.
\newblock In \emph{{AIOPS 2020 - International Workshop on Artificial
  Intelligence for IT Operations}}, Dubai, United Arab Emirates, December 2020.
\newblock URL \url{https://hal.inria.fr/hal-02948735}.

\bibitem[Wu and Keogh(2021)]{Wu_2021}
Renjie Wu and Eamonn Keogh.
\newblock Current time series anomaly detection benchmarks are flawed and are
  creating the illusion of progress.
\newblock \emph{IEEE Transactions on Knowledge and Data Engineering}, page
  1–1, 2021.
\newblock ISSN 2326-3865.
\newblock \doi{10.1109/tkde.2021.3112126}.
\newblock URL \url{http://dx.doi.org/10.1109/TKDE.2021.3112126}.

\bibitem[Xie et~al.(2018)Xie, Dai, Du, Hovy, and Neubig]{xie2018controllable}
Qizhe Xie, Zihang Dai, Yulun Du, Eduard Hovy, and Graham Neubig.
\newblock Controllable invariance through adversarial feature learning.
\newblock In \emph{Proceedings of the 31st International Conference on Neural
  Information Processing Systems}. Curran Associates Inc., 2018.

\bibitem[Zerveas et~al.(2021)Zerveas, Jayaraman, Patel, Bhamidipaty, and
  Eickhoff]{transformerReps}
George Zerveas, Srideepika Jayaraman, Dhaval Patel, Anuradha Bhamidipaty, and
  Carsten Eickhoff.
\newblock A transformer-based framework for multivariate time series
  representation learning.
\newblock In \emph{Proceedings of the 27th ACM SIGKDD Conference on Knowledge
  Discovery and Data Mining}, KDD '21, page 2114–2124, New York, NY, USA,
  2021. Association for Computing Machinery.
\newblock ISBN 9781450383325.
\newblock \doi{10.1145/3447548.3467401}.
\newblock URL \url{https://doi.org/10.1145/3447548.3467401}.

\end{thebibliography}
\end{document}